\documentclass{article}

\usepackage[pdftex]{graphicx}
\usepackage[table,dvipsnames]{xcolor}
\usepackage[round]{natbib}
\usepackage[export]{adjustbox}[2011/08/13]

\usepackage{microtype}
\usepackage{subfigure}
\usepackage{booktabs} 

\usepackage{amsmath}
\usepackage{amssymb}
\usepackage{bm}
\usepackage{multirow}
\usepackage{makecell} 

\usepackage[algo2e,ruled]{algorithm2e}

\usepackage[colorlinks]{hyperref}

\hypersetup{citecolor=[rgb]{0.45,0.0,1.0},linkcolor=Periwinkle}

\usepackage{hyperref}

\usepackage[accepted]{icml2020}

\usepackage{bbold}

\newcommand{\bff}{\mathbf{f}}
\newcommand{\bu}{\mathbf{u}}
\newcommand{\by}{\mathbf{y}}
\newcommand{\bx}{\mathbf{x}}
\newcommand{\bX}{\mathbf{X}}
\newcommand{\bz}{\mathbf{z}}
\newcommand{\bk}{\mathbf{k}}
\newcommand{\bK}{\mathbf{K}}
\newcommand{\bKt}{\mathbf{\tilde{K}}}
\newcommand{\bL}{\mathbf{L}}
\newcommand{\bZ}{\mathbf{Z}}

\newcommand{\EE}{\mathbb{E}}

\newcommand{\NN}{\mathcal{N}}
\newcommand{\LL}{\mathcal{L}}
\newcommand{\qtilde}{\tilde{q}}
\newcommand{\KL}{{\rm KL}}
\newcommand{\bmm}{\bm{m}}
\newcommand{\bS}{\mathbf{S}}
\newcommand{\bhalfK}{\mathbf{\Lambda}}
\newcommand{\OO}{\mathcal{O}}

\newcommand{\sigmaobs}{\sigma_{\rm obs}}

\newcommand{\betareg}{\beta_{\rm reg}}
\newcommand{\thetaker}{\theta_{\rm ker}}

\icmltitlerunning{Parametric Gaussian Process Regressors}

\begin{document}

\twocolumn[
\icmltitle{Parametric Gaussian Process Regressors}




\begin{icmlauthorlist}
\icmlauthor{Martin Jankowiak}{broad}
\icmlauthor{Geoff Pleiss}{corn}
\icmlauthor{Jacob R. Gardner}{upenn}
\end{icmlauthorlist}

\icmlaffiliation{broad}{The Broad Institute, Cambridge, MA, USA}
\icmlaffiliation{corn}{Dept. of Computer Science, Cornell University, Ithaca, NY, USA}
\icmlaffiliation{upenn}{Dept. of Computer and Information Science, University of Pennsylvania, Philadelphia, PA, USA }

\icmlcorrespondingauthor{Martin Jankowiak}{mjankowi@broadinstitute.org}

\icmlkeywords{Machine Learning, ICML, gaussian processes}

\vskip 0.3in
]



\printAffiliationsAndNotice{} 

\begin{abstract}
The combination of inducing point methods with stochastic variational inference has enabled approximate
Gaussian Process (GP) inference on large datasets. Unfortunately, the resulting predictive distributions often exhibit substantially underestimated uncertainties.
Notably, in the regression case the predictive variance is typically dominated by observation noise, yielding uncertainty estimates that make little use of the input-dependent function uncertainty that makes GP priors attractive.
In this work we propose two simple methods for scalable GP regression that address this issue and thus yield substantially improved predictive uncertainties.
The first applies variational inference to FITC (Fully Independent Training Conditional; Snelson et.~al.~2006).
The second bypasses posterior approximations and instead directly targets the posterior predictive distribution.
In an extensive empirical comparison with a number of alternative methods for scalable GP regression,
we find that the resulting predictive distributions exhibit significantly better calibrated uncertainties and higher log likelihoods---often by as much as half a nat per datapoint.
\end{abstract}

\section{Introduction}
\label{sec:intro}

\begin{figure}[t]
  \centering
  {\includegraphics[width=\linewidth]{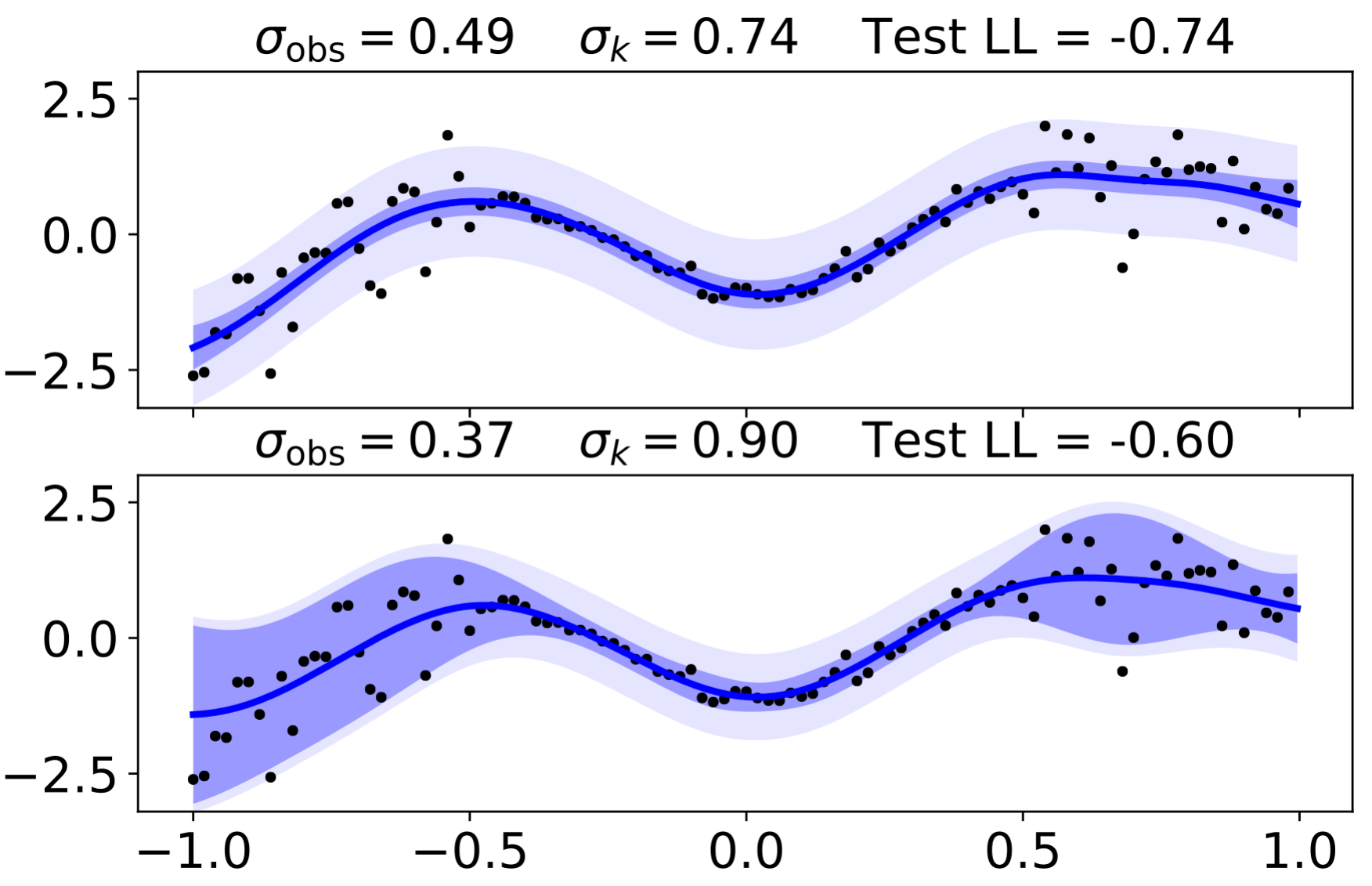}}
  \caption{We depict GP regressors fit to a heteroscedastic dataset using two different inference algorithms.
  Solid lines depict mean predictions and 2-$\sigma$ uncertainty bands are in blue.
  In the lower panel, fit with the PPGPR approach described in Sec.~\ref{sec:mlpred},
  significant use is made of input-dependent function uncertainty (dark blue),
  while in the upper panel, fit with variational inference (see Sec.~\ref{sec:svgp}), the predictive uncertainty is dominated by the observation noise $\sigmaobs^2$ (light blue) and
  the kernel scale $\sigma_k$ is smaller.}
  \label{fig:univariate}
\end{figure}

Machine learning is finding increasing use in applications where autonomous decisions are driven by predictive models.
For example, machine learning can be used to steer dynamic load balancing in critical electrical systems,
and autonomous vehicles use machine learning algorithms to detect and
classify objects in unpredictable weather conditions and decide whether to brake.
As machine learning models increasingly become deployed as components in larger decision making pipelines,
it is essential that models be able to reason about uncertainty and risk. Techniques drawn from probabilistic machine learning offer the ability to deal with these challenges by offering predictive models with simple and interpretable probabilistic outputs.

Recent years have seen extensive use of variational inference \citep{jordan1999introduction} as a workhorse inference algorithm for a
variety of probabilistic models.  The popularity of variational inference has been been driven by a number of different factors, including:
i) its amenability to data subsampling \citep{hoffman2013stochastic};
ii) its applicability to black-box non-conjugate models \citep{kingma2013auto, rezende2014stochastic};
and iii) its suitability for GPU acceleration.

The many practical successes of variational inference notwithstanding, it has long been recognized that variational inference
often results in overconfident uncertainty estimates.\footnote{For example see \citet{turner+sahani:2011a} for a discussion of this point in the context of time series models.}
This problem can be especially acute for Gaussian Process (GP) models \citep{bauer2016understanding}. In particular GP regressors
fit with variational inference tend to apportion most of the predictive variance to the input-independent observation noise,
making little use of the input-dependent function uncertainty that makes GP priors attractive in the first place (see Fig.~\ref{fig:ucivarratio}
in Sec.~\ref{sec:exp} for empirical evidence).
As we explain in Sec.~\ref{sec:preddisc}, this tendency can be understood as resulting from the asymmetry with which variational inference---through its reliance on Jensen's inequality---treats the various contributions to the uncertainty
in output space, in particular its asymmetric treatment of the observation noise.

In this work we propose two simple solutions that correct this undesirable behavior.
In the first we apply variational inference to the well known FITC (Fully Independent Training Conditional; \citet{snelson2006sparse})
method for sparse GP regression.
In the second we directly target the predictive distribution---bypassing posterior approximations entirely---to formulate an objective function that treats the various contributions to predictive variance on an equal footing.
As we show empirically in Sec.~\ref{sec:exp}, the predictive distributions resulting from these two parametric approaches to GP regression
exhibit better calibrated uncertainties and higher log likelihoods than those obtained with existing methods for scalable GP regression.

\section{Background}
\label{sec:background}

This section is organized as follows.
In Sec.~\ref{sec:gpr}-\ref{sec:sparse} we review the basics of Gaussian Processes and inducing point methods.
In Sec.~\ref{sec:inference} we review various approaches to scalable GP inference that will serve as the baselines in
our experiments.
In Sec.~\ref{sec:preddist} we describe the predictive distributions that result from these methods.
Finally in Sec.~\ref{sec:fitc} we review FITC \cite{snelson2006sparse}---an approach to sparse GP regression
that achieves scalability by suitably modifying the regression model---as it will serve
as one of the ingredients to the approach introduced in Sec.~\ref{sec:vfitc}.
We also use this section to establish our notation.

\subsection{Gaussian Process Regression}
\label{sec:gpr}

In probabilistic modeling Gaussian Processes offer powerful non-parametric function priors that are useful in a variety of regression and classification tasks \citep{rasmussen2003gaussian}. For a given input space $\mathbb{R}^d$
GPs are entirely specified by a covariance function or kernel $k: \mathbb{R}^d \times \mathbb{R}^d \to \mathbb{R}$
and a mean function $\mu: \mathbb{R}^d \to \mathbb{R}$. Different choices of $\mu$ and $k$ allow the modeler to
encode prior information about the generative process.
In the prototypical case of univariate regression the joint density takes the form\footnote{In the following we will
suppress dependence on the kernel $k$ and the mean function $\mu$.}
\begin{equation}
\label{eqn:unireg}
p(\by, \bff | \bX) = p(\by|\bff, \sigmaobs^2) p(\bff | \bX, k, \mu)
\end{equation}
where $\by$ are the real-valued targets, $\bff$ are the latent function values, $\bX = \{ \bx_i \}_{i=1}^N$ are the $N$ inputs with $\bx_i \in \mathbb{R}^d$,
and $\sigmaobs^2$ is the variance of the Normal likelihood $p(\by|\cdot)$.
The marginal likelihood takes the form
\begin{equation}\label{eqn:marg}
p(\by|\bX) = \int \! d \bff \; p(\by|\bff, \sigmaobs^2) p(\bff | \bX)
\end{equation}
Eqn.~\ref{eqn:marg} can be computed analytically, but doing so is computationally prohibitive for large datasets.
This is because the cost scales as $\OO(N^3)$ from the terms involving $\bK_{NN}^{-1}$ and ${\rm log  det \;} \bK_{NN}$ in
Eqn.~\ref{eqn:marg}, where  $\bK_{NN} = k(\bX,\bX)$. This necessitates approximate methods when $N$ is large.

\subsection{Sparse Gaussian Processes}
\label{sec:sparse}

Over the past two decades significant progress has been made in scaling Gaussian Process inference to large datasets.
The key technical innovation was the development of inducing point methods \citep{snelson2006sparse,titsias2009variational, hensman2013gaussian}, which we now review. By introducing inducing
variables $\bu$ that depend on
variational parameters $\{ \bz_m \}_{m=1}^{M}$, where $M={\rm dim}(\bu) \ll N$ and with each $\bz_m \in \mathbb{R}^d$, we augment the GP prior as follows:
\begin{equation}\nonumber
p(\bff|\bX) \rightarrow  p(\bff|\bu,\bX,\bZ) p(\bu|\bZ)
\end{equation}
We then appeal to Jensen's inequality and lower bound the log joint density over the targets and inducing variables:
\begin{align}
\label{eqn:jensenenergy}
\begin{aligned}
\log p(\by, \bu |\bX, \bZ) &= \log \int d\bff p(\by|\bff) p(\bff|\bu) p(\bu)  \\
&\ge  \EE_{p(\bff|\bu)} \left[ \log p(\by|\bff)  +\log p(\bu) \right] \\
 &=  \! \sum_{i=1}^N \! \log \mathcal{N}(y_i | \bk_i^{T} \bK_{MM}^{-1} \bu, \sigmaobs^2)  \\ 
 &- \tfrac{1}{2\sigmaobs^2}  {\rm Tr} \;\! \bKt_{NN} + \log p(\bu)
 \end{aligned}
\end{align}
where $\bk_i = k(\bx_i, \bZ)$, $\bK_{MM}=k(\bZ,\bZ)$ and $\bKt_{NN}$ is given by
\begin{equation}
\bKt_{NN} = \bK_{NN} - \bK_{NM}  \bK_{MM} ^{-1} \bK_{MN}
\end{equation}
with $\bK_{NM} =  \bK_{MN}^{\rm T} = k(\bX,\bZ)$. The essential characteristics of Eqn.~\ref{eqn:jensenenergy} are that:
i) it replaces expensive computations involving $\bK_{NN}$ 
with cheaper computations like $\bK_{MM}^{-1}$ that scale as $\OO(M^3)$; and ii) it is amenable to data subsampling,
since the log likelihood and trace terms factorize as sums over datapoints $(y_i, \bx_i)$.

\subsection{Approximate GP Inference}
\label{sec:inference}

We now describe how Eqn.~\ref{eqn:jensenenergy}
can be used to construct a variety of algorithms for scalable GP inference.
We limit ourselves to algorithms that satisfy two desiderata:
i) support for data subsampling;\footnote{For this reason we do not consider the method in \citep{titsias2009variational}.
Note that all the inference algorithms that we describe that make use of Eqn.~\ref{eqn:jensenenergy} automatically
inherit its support for data subsampling.} and
ii) the result of inference is a compact artifact that enables fast predictions at test time.\footnote{Consequently
we do not consider MCMC algorithms like the Stochastic gradient HMC algorithm explored in the context of
deep gaussian processes in \citep{havasi2018inference}, and which also utilizes
Eqn.~\ref{eqn:jensenenergy}.} The rest of this section is organized as follows.
In Sec.~\ref{sec:svgp} we describe SVGP \citep{hensman2013gaussian}---currently the most popular method for
scalable GP inference. In Sec.~\ref{sec:map} we describe how Eqn.~\ref{eqn:jensenenergy} can be leveraged
in the context of MAP (maximum a posteriori) inference.
Then in Sec.~\ref{sec:robust} we briefly review how ideas from robust variational inference can be applied to the GP setting.

\subsubsection{SVGP}
\label{sec:svgp}

SVGP proceeds by introducing a multivariate Normal variational distribution
$q(\bu) = \NN(\bmm, \bS)$.
The parameters $\bmm$ and $\bS$ are optimized using the ELBO (evidence lower bound), which is the expectation
of Eqn.~\ref{eqn:jensenenergy} w.r.t.~$q(\bu)$ plus an entropy term term $H[q(\bu)]$:
\begin{align}
\begin{aligned}
\label{eqn:svgp}
&\mathcal{L}_{\rm svgp}  = \EE_{q(\bu)} \left[ \log p(\by, \bu |\bX, \bZ) \right] + H[q(\bu)] \\
&\phantom{\mathcal{L}_{\rm svgp}}= \sum_{i=1}^N \log   \mathcal{N}(y_i | \bk_i^{T} \bK_{MM}^{-1} \bmm, \sigmaobs^2)
 - \tfrac{1}{2\sigmaobs^2} {\rm Tr}\; \bKt_{NN} \\
&\phantom{\mathcal{L}_{\rm svgp}}- \tfrac{1}{2\sigmaobs^2}  \sum_{i=1}^N  \bk_i^{T} \bK_{MM}^{-1} \bS \bK_{MM}^{-1}  \bk_i
-  \KL(q(\bu) | p(\bu))
\end{aligned}
\raisetag{83pt}
\end{align}
where KL denotes the Kullback-Leibler divergence. Eqn.~\ref{eqn:svgp} can be rewritten more compactly as
\begin{align}
\begin{aligned}
\label{eqn:svgp2}
\mathcal{L}_{\rm svgp}  =& \sum_{i=1}^N \left\{ \log   \mathcal{N}(y_i | \mu_\bff(\bx_i), \sigmaobs^2)
 - \frac{\sigma_\bff(\bx_i)^2}{2\sigmaobs^2} \right\}  \\
& -  \KL(q(\bu) | p(\bu))
\end{aligned}
\raisetag{13pt}
\end{align}
where $\mu_\bff(\bx_i)$ is the predictive mean function given by
\begin{equation}
\label{eqn:meanfunc}
\mu_\bff(\bx_i) = \bk_i^{T} \bK_{MM}^{-1} \bmm
\end{equation}
and where $\sigma_\bff(\bx_i)^2 \equiv \rm{Var}[f_i | \bx_i] $ denotes the latent function variance
\begin{equation}
\label{eqn:fvar}
\sigma_\bff(\bx_i)^2 =  \bKt_{ii} + \bk_i^{T}  \bK_{MM}^{-1} \bS \bK_{MM}^{-1}  \bk_i
\end{equation}
$\mathcal{L}_{\rm svgp}$, which depends on $\bmm, \bS, \bZ, \sigmaobs$ and the various kernel hyperparameters $\thetaker$,
can then be maximized with gradient methods. Below we refer to the resulting GP regression method as {\bf SVGP}.
Note that throughout this work we consider a variant of SVGP in which the KL divergence term in Eqn.~\ref{eqn:svgp2} is
scaled by a positive constant $\betareg > 0$.

\subsubsection{MAP}
\label{sec:map}

In contrast to SVGP, which maintains a distribution over the inducing variables $\bu$, MAP is a particle method
in which we directly optimize a single point $\bu \in \mathbb{R}^M$ rather than a distribution over $\bu$.
In particular we simply maximize Eqn.~\ref{eqn:jensenenergy} evaluated at $\bu$.
Note that the term $\log p(\bu)$ serves as a regularizer.
In the following we refer to this inference procedure as {\bf MAP}.
To the best of our knowledge, it has not been considered before in the sparse GP literature.

\subsubsection{Robust Gaussian Processes}
\label{sec:robust}

\citet{knoblauch2019generalized,knoblauch2019dgp} consider modifications to the typical variational objective (e.g.~Eqn.~\ref{eqn:svgp}), which consists of an expected log likelihood and a KL divergence term.
In particular, they replace the expected log likelihood loss with an alternative divergence like the gamma divergence.
This divergence raises the likelihood to a power
\begin{equation}
\label{eqn:gammarobust}
\log p(\by|\bff) \rightarrow p(\by|\bff)^{\gamma-1}
\end{equation}
where typically $\gamma \in (1.0, 1.1)$.\footnote{See \citep{cichocki2010families} for a detailed discussion of this and other divergences.}
Empirically \citet{knoblauch2019dgp} demonstrates that this modification can yield better performance than SVGP on regression tasks.
We refer to this inference procedure as $\bm{\gamma}$-{\bf Robust}.

\subsection{Predictive Distributions}
\label{sec:preddist}

All of the methods in Sec.~\ref{sec:inference} yield predictive distributions of the same form. In particular,
conditioned on the inducing variable $\bu$ the predictive distribution at input $\bx^*$ is given by
\begin{equation}
\label{eqn:preddist}
p(\by^* | \bx^*, \bu) = \NN( \by^* | \bk_*^{T} \bK_{MM}^{-1} \bu, \bKt_{**} + \sigmaobs^2)
\end{equation}
Integrating out $\bu \sim \NN(\bmm, \bS)$ then yields
\begin{equation}
\label{eqn:preddist2}
p(\by^* | \bx^*) = \NN( \by^* |  \mu_\bff(\bx^*), \sigma_\bff(\bx^*)^2 + \sigmaobs^2)
\end{equation}
where $\mu_\bff(\bx^*)$ is the predictive mean function in Eqn.~\ref{eqn:meanfunc} and
$\sigma_\bff(\bx^*)^2$ is the latent function variance in Eqn.~\ref{eqn:fvar}. Note that
since MAP can be viewed as a degenerate limit of SVGP, the predictive distribution for MAP can be obtained by taking the
limit $\bS \rightarrow 0$ in $\sigma_\bff(\bx^*)^2$.

\subsection{FITC}
\label{sec:fitc}

FITC (Fully Independent Training Conditional) \cite{snelson2006sparse} is
a method for sparse GP regression that is formulated using a joint probability
\begin{equation}
\label{eqn:fitc}
p(\by , \bu) = \prod_{i=1}^N p(y_i | \bk_i^{T} \bK_{MM}^{-1} \bu, \bKt_{ii} + \sigmaobs^2) p(\bu )
\end{equation}
with a modified likelihood corresponding to the conditional predictive distribution in Eqn.~\ref{eqn:preddist} (for additional interpretations see \citet{bauer2016understanding}).
As noted by \citet{snelson2006sparse}
this can be viewed as a standard regression model with a particular form of parametric
mean function and input-dependent observation noise.
Integrating out the latent function values $\bu$ results in a marginal likelihood
\begin{equation}
\label{eqn:fitcmarg}
p(\by) = \NN(\by | \bm{0}, \bK_{NM} \bK_{MM}^{-1} \bK_{MN} + {\rm diag}(\bKt_{NN}) + \sigmaobs^2 \mathbb{1})
\end{equation}
that can be used for training with $\mathcal{O}(NM^2 + M^3)$ computational complexity per gradient step. Note that the structure of the covariance matrix in Eqn.~\ref{eqn:fitcmarg} prevents mini-batch training, limiting FITC to moderately large datasets.

\section{Parametric Gaussian Process Regressors}
\label{sec:paragp}

\SetAlCapHSkip{0em}
\setlength\algomargin{0em}
\begin{algorithm2e}[t!]
  \SetKwInOut{Input}{Input}
  \SetKwInOut{Output}{Output}
  \newlength\inputlen
  \newcommand\NextInput[1]{%
    \settowidth\inputlen{\KwIn{}}%
    \setlength\hangindent{1.5\inputlen}%
    \hspace*{\inputlen} #1\\
  }
  \newcommand\graycomment[1]{\footnotesize\ttfamily\textcolor{gray}{#1}}
  \SetCommentSty{graycomment}
  \SetKw{Break}{break}
  \SetKwData{tol}{tolerance}
  \SetKwFunction{optim}{optim}
  \parbox{\linewidth}{\caption{Scalable GP Regression. All of the inference algorithms
  we consider follow the same basic pattern and only differ in the form
  of the objective function, e.g.~$\LL_{\rm svgp}$ (Eqn.~\ref{eqn:svgp2}), $\LL_{\rm vfitc}$
  (Eqn.~\ref{eqn:vfitcelbo}) or $\LL_{\rm ppgpr}$ (Eqn.~\ref{eqn:regpred2}).
  Similarly for all methods the predictive distribution is given by Eqn.~\ref{eqn:preddist2}.
    See Sec.~\ref{sec:complexity} in the supplementary materials for a discussion of the time and space complexity of each method.}}
  \label{alg:gp}
    \KwIn{$\mathcal L$ -- objective function}
    \NextInput{\optim{} -- gradient-based optimizer}
    \NextInput{$\mathcal{D} = \{\bx_i, y_i\}_{i=1}^N$ -- training data}
    \NextInput{$\bmm, \bS, \bZ, \thetaker$ -- initial parameters}
    \KwOut{$\bmm, \bS, \bZ, \thetaker$.}
    \BlankLine
    \While{\text{not converged}}{
        Choose a random mini-batch $\mathcal{D}_{\rm mb} \subset \mathcal{D}$ \\
        Form an unbiased estimate $\hat{\mathcal{L}}(\mathcal{D}_{\rm mb})$ \\
      Gradient step: $\bmm, \bS, \bZ, \thetaker$ $\gets$ $\optim \left( \hat{\mathcal{L}}(\mathcal{D}_{\rm mb}) \right)$
    }
\end{algorithm2e}

Before introducing the two scalable methods for GP regression we consider in this work, we examine one of the salient
characteristics of the SVGP objective Eqn.~\ref{eqn:svgp2} referred to in the introduction.
As discussed in Sec.~\ref{sec:preddist}, in SVGP the predictive variance ${\rm Var}[y^* | \bx^*]$ at an input $\bx^*$ has two contributions,
one that is input-dependent, i.e.~$\sigma_\bff(\bx^*)^2$, 
and one that is input-independent, i.e.~$\sigmaobs^2$:
\begin{equation}
\begin{split}
{\rm Var}[y^* | \bx^*] &= \sigmaobs^2 + \sigma_\bff(\bx^*)^2 \\
                                 &= \sigmaobs^2 + \bKt_{**} + \bk_*^{T}  \bK_{MM}^{-1} \bS \bK_{MM}^{-1}  \bk_*
\end{split}
\end{equation}
These two contributions appear \emph{asymmetrically} in  Eqn.~\ref{eqn:svgp2}; in particular $\sigma_\bff(\bx_i)^2$ does not appear
in the data fit term that results from the expected log likelihood $\EE_{q(\bu)} \left[ \log p(\by_i |\bx_i,  \bu) \right] $. We expect
this behavior to be undesirable, since it leads to a mismatch between the training objective and the predictive distribution used at test time.

In the following we introduce two approaches that address this asymmetry.
Crucially, unlike the inference strategies
outlined in Sec.~\ref{sec:inference}, neither approach makes use of the lower-bound energy surface
in Eqn.~\ref{eqn:jensenenergy}.
As we will see, the first approach, Variational FITC, introduced
in Sec.~\ref{sec:vfitc}, only partially removes the asymmetry, while the second Parametric Predictive GP approach, introduced
in Sec.~\ref{sec:mlpred}, restores full symmetry between the training objective and the test time predictive distribution.
For more discussion of this point see Sec.~\ref{sec:preddisc}.

\subsection{Variational FITC}
\label{sec:vfitc}

Variational FITC proceeds by applying variational inference to the FITC model defined in Eqn.~\ref{eqn:fitc}.
That is, we introduce a multivariate Normal variational distribution $q(\bu) = \NN(\bmm, \bS)$ and compute the ELBO:
\begin{align}
\begin{aligned}
\label{eqn:vfitcelbo}
&\mathcal{L}_{\rm vfitc}  = \EE_{q(\bu)} \left[ \log p(\by, \bu |\bX, \bZ) \right] + H[q(\bu)] \\
&\phantom{\mathcal{L}_{\rm svgp}}= \sum_{i=1}^N \log   \mathcal{N}(y_i | \mu_\bff(\bx_i), \bKt_{ii} + \sigmaobs^2) \\
&\phantom{\mathcal{L}_{\rm svgp}}- \tfrac{1}{2}  \sum_{i=1}^N \frac{ \bk_i^{T} \bK_{MM}^{-1} \bS \bK_{MM}^{-1}  \bk_i } {\bKt_{ii} + \sigmaobs^2}
-  \KL(q(\bu) | p(\bu))
\raisetag{57pt}
\end{aligned}
\end{align}
The parameters $\bmm$, $\bS$, $\bZ$, as well as the observation noise $\sigmaobs$ and kernel hyperparameters $\thetaker$ can then
be optimized by maximizing Eqn.~\ref{eqn:vfitcelbo} using gradient methods (see Algorithm~\ref{alg:gp}).
Note that since the inducing point locations $\bZ$ appear in the model, this is properly understood as a parametric model.

We note that, unlike FITC, the objective in Eqn.~\ref{eqn:vfitcelbo} readily supports mini-batch training, and is thus suitable
for very large datasets.
Below we refer to this method as {\bf VFITC}.

\subsection{Parametric Predictive GP Regression}
\label{sec:mlpred}

As we discuss further in the next section, using the VFITC objective only partially addresses the 
asymmetric treatment of $\sigmaobs^2$ and $\sigma_\bff(\bx^*)^2$ in Eqn.~\ref{eqn:svgp2}. 
We now describe our second approach to scalable GP regression, which is motivated by the goal of 
restoring full symmetry between the training objective and the test time predictive distribution.
To accomplish this, we introduce a parametric GP regression model formulated directly
in terms of the family of predictive distributions given in Eqn.~\ref{eqn:preddist2}.
We then introduce an objective function based on the KL divergence between $p(y | \bx)$ and $p_{\rm data}(y | \bx)$
\begin{equation}
\begin{split}
\label{eqn:unregpred}
\LL_{\rm ppgpr}^\prime &= - \EE_{p_{\rm data}(\bx)} \; \KL( p_{\rm data}(y|\bx) || p(y|\bx)) \\
 &\rightarrow \EE_{p_{\rm data}(y, \bx)} \left[ \log p(y|\bx) \right]
\end{split}
\end{equation}
where $p_{\rm data}(y, \bx)$ is the empirical distribution over training data. In the second line we have dropped the entropy term $-\EE_{p_{\rm data}(y|\bx)} \left[ \log p_{\rm data}(y|\bx)\right]$, since it is a constant w.r.t.~to the maximization
problem.
We obtain our final objective function by adding an optional regularization term modulated by a positive constant $\betareg > 0$:
\begin{equation}
\label{eqn:regpred}
\LL_{\rm ppgpr} =  \EE_{p_{\rm data}(y, \bx)} \left[ \log p(y|\bx) \right] - \betareg \KL(q(\bu) || p(\bu))
\end{equation}
Note that apart from the regularization term, maximizing this objective function corresponds to doing maximum likelihood estimation
of a parametric model defined by $p(y|\bx)$.
The objective in Eqn.~\ref{eqn:regpred} can be expanded as
\begin{align}
\label{eqn:regpred2}
\begin{aligned}
\LL_{\rm ppgpr} = & \sum_{i=1}^N \log  \mathcal{N}(y_i | \mu_\bff(\bx_i), \sigmaobs^2 + \sigma_\bff(\bx_i)^2) \\
&- \betareg \KL(q(\bu) || p(\bu))
\raisetag{12pt}
\end{aligned}
\end{align}
where $\sigma_\bff(\bx_i)^2 = \rm{Var}[f_i | \bx_i]$ is the function variance defined in Eqn.~\ref{eqn:fvar}.
The parameters $\bmm$, $\bS$, $\bZ$, as well as the observation noise $\sigmaobs$ and kernel hyperparameters $\thetaker$ can then
be optimized by maximizing Eqn.~\ref{eqn:regpred} using gradient methods (see Algorithm~\ref{alg:gp}).
We refer to this model class as {\bf PPGPR}.

When $\betareg=1$ the form of the objective in Eqn.~\ref{eqn:regpred} can be motivated by a connection to
Expectation Propagation \citep{minka2004power}; see Sec.~\ref{sec:ep}
in the supplementary materials and \citep{li2017dropout} for further discussion.\footnote{We thank Thang Bui for pointing out this connection.}

\subsection{Discussion}
\label{sec:preddisc}

What are the implications of employing the scalable GP regressors described in Sec.~\ref{sec:vfitc} and Sec.~\ref{sec:mlpred}?\footnote{See Sec~\ref{sec:additional} in the supplementary materials for a comparison to exact GPs.}
It is helpful to compare the objective functions in Eqn.~\ref{eqn:vfitcelbo} and Eqn.~\ref{eqn:regpred} to the
SVGP objective in Eqn.~\ref{eqn:svgp2}.
In Eqn.~\ref{eqn:vfitcelbo} we obtain a data fit term
\begin{equation}
\label{eqn:vfitcdatafit}
\LL_{\rm vfitc} \supset -\tfrac{1}{2}\tfrac{1}{\sigmaobs^2 + \bKt_{ii}}  \left \vert y_i - \mu_\bff(\bx_i) \right\vert^2
\end{equation}
while in Eqn.~\ref{eqn:regpred2} we obtain a data fit term
\begin{equation}
\label{eqn:preddatafit}
\LL_{\rm ppgpr} \supset -\tfrac{1}{2}\tfrac{1}{\sigmaobs^2 + \sigma_\bff(\bx_i)^2}  \left \vert y_i - \mu_\bff(\bx_i) \right\vert^2
\end{equation}
In contrast in SVGP the corresponding data fit term is
\begin{equation}
\label{eqn:svgpdatafit}
\LL_{\rm svgp} \supset -\tfrac{1}{2}\tfrac{1}{\sigmaobs^2}  \left \vert y_i - \mu_\bff(\bx_i)  \right\vert^2
\end{equation}
with $\sigmaobs^2$ in the denominator.
We reiterate that in all three approaches the predictive variance ${\rm Var}[y^* | \bx^*]$ at an input $\bx^*$ is given by the formula
\begin{equation}
\label{eqn:predvar}
\begin{split}
&{\rm Var}[y^* | \bx^*] = \sigmaobs^2 + \sigma_\bff(\bx^*)^2
\end{split}
\end{equation}
where $ \sigma_\bff(\bx^*)^2$ is the latent function variance at $\bx^*$ (see Eqn.~\ref{eqn:fvar}).
Thus SVGP---and, more generally, variational inference for any regression model with a Normal likelihood---makes
an arbitrary\footnote{Arbitrary from the point of view of output space. Depending on the particular application and structure of the model, distinctions between different contributions to the predictive variance may be of interest.} distinction between the observation variance $\sigmaobs^2$ and the latent function variance $\sigma_\bff(\bx^*)^2$,
even though both terms contribute symmetrically to the total predictive variance in Eqn.~\ref{eqn:predvar}. Moreover,
this asymmetry will be inherited by any method that makes use of Eqn.~\ref{eqn:jensenenergy}, e.g.~the MAP
procedure described in Sec.~\ref{sec:map}.

When the priority is predictive performance---the typical case in machine learning---this
asymmetric treatment of the two contributions to the predictive variance is troubling.
As we will see in experiments (Sec.~\ref{sec:exp}), the difference between Eqn.~\ref{eqn:preddatafit} and Eqn.~\ref{eqn:svgpdatafit}
has dramatic consequences. In particular the data fit term in SVGP does nothing to encourage function
variance $\sigma_\bff(\bx^*)$.
As a consequence for many datasets $\sigma_\bff(\bx^*) \ll \sigmaobs$; i.e.~\emph{most
of the predictive variance is explained by the input-independent observation noise}.

In contrast, the data fit term in Eqn.~\ref{eqn:preddatafit}
directly encourages large $\sigma_\bff(\bx^*)$, typically resulting in behavior opposite to that of SVGP, i.e.~$\sigma_\bff(\bx^*) \gg \sigmaobs$.
This is gratifying because---after having gone to the effort to introduce an input-dependent kernel and learn an appropriate
geometry on the input space---we end up with predictive variances that make substantial use of the input-dependent kernel.

The case of Variational FITC (see Eqn.~\ref{eqn:vfitcdatafit}) is intermediate in that the objective function directly incentivizes
non-negligible latent function variance through the term $\bKt_{ii}(\bx_i)$, while the $\bS$-dependent contribution to $\sigma_\bff(\bx_i)$ is
still treated asymmetrically (since it does not appear in the data fit term).

As argued by \citep{bauer2016understanding}, methods based on FITC---and by extension our Parametric Predictive GP approach---tend to
\emph{underestimate} the observation noise $\sigmaobs^2$, while variational methods like SVGP or the closely related method introduced
by \cite{titsias2009variational} tend to \emph{overestimate} $\sigmaobs^2$.
While the possibility of underestimating $\sigmaobs^2$  is indeed a concern for our approach,
our empirical results in Sec.~\ref{sec:exp} suggest that, this tendency notwithstanding, our methods yield excellent predictive performance.

\begin{figure*}[t!]
  \centering
  \includegraphics[width=1.0\textwidth,center]{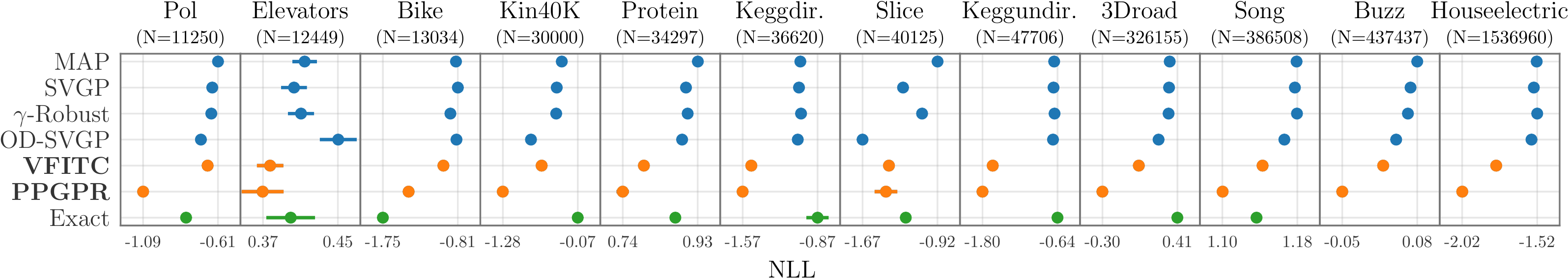}
    \caption{We depict test negative log likelihoods (NLL) for twelve univariate regression datasets (lower is better). Results are averaged over ten random train/test/validation splits. See Sec.~\ref{sec:uci} for details. Here and elsewhere horizontal error bars depict standard errors.}
  \label{fig:ucill}
\end{figure*}

\begin{figure*}[t!]
  \centering
  \includegraphics[width=1.0\textwidth,center]{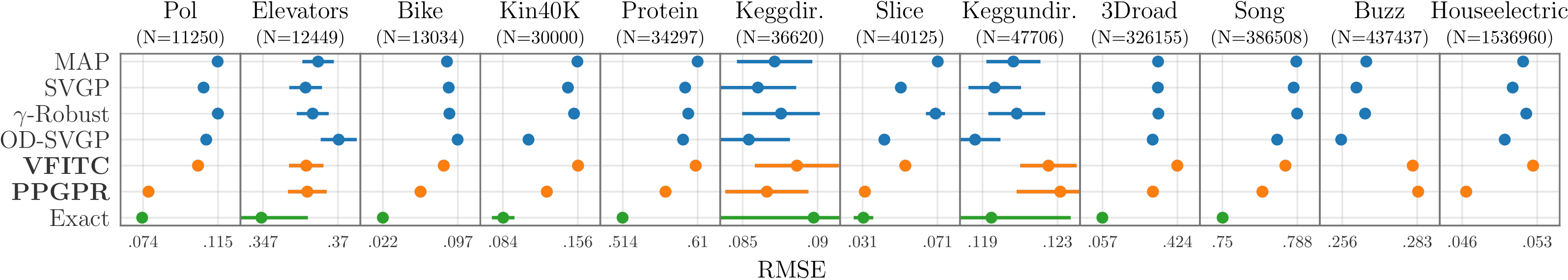}
    \caption{We depict test root mean squared errors (RMSE) for twelve univariate regression datasets (lower is better). Results are averaged over ten random train/test/validation splits. See Sec.~\ref{sec:uci} for details.}
  \label{fig:ucirmse}
\end{figure*}

\subsection{Additional Variants}
\label{sec:variants}

A number of variants to the Parametric Predictive GP approach outlined in Sec.~\ref{sec:mlpred}  immediately suggest themselves. One possibility is to take
the formal limit $\bS \rightarrow 0$ in $q(\bu)$. In this limit $q(\bu)$ is a Dirac delta distribution,
the function variance $\sigma_\bff(\bx_i)^2 \rightarrow \bKt_{ii}$, and the number of parameters
is now linear in $M$ instead of quadratic.\footnote{Additionally in the regularizer we make the replacement
$-\betareg \KL(q(\bu) | p(\bu)) \rightarrow \betareg \log p(\bu)$.} Below we refer to this variant as {\bf PPGPR}-$\bm{\delta}$.
Another possibility is to restrict the covariance matrix $\bS$ in $q(\bu)$ to be diagonal; we refer to this `mean field'
variant as {\bf PPGPR-MF}. Yet another possibility is to decouple the parametric mean function
$\mu_\bff(\bx)$ and variance function $\sigma_\bff(\bx)^2$ used to define $p(y|\bx)$ in Eqn.~\ref{eqn:regpred}, i.e.~use
 separate inducing point locations $\bZ_\mu$ and $\bZ_\sigma$ for additional flexibility. This is conceptually similar
to the decoupled approach introduced in \citet{cheng2017variational}, with the difference that our parametric
modeling approach is not constrained by the need to construct a well-defined variational inference problem in an infinite-dimensional
RKHS. Below we refer to this decoupled approach together with a diagonal covariance $\bS$ as {\bf PPGPR-MFD}.
We refer to the variant of PPGPR that is closest to SVGP (because it utilizes a full-rank covariance matrix $\bS=\bL \bL^{\rm T}$) as
{\bf PPGPR-Chol}.

A number of other variants are also possible. For example we might replace the KL regularizer
in Eqn.~\ref{eqn:regpred} with another divergence, for example a R{\'e}nyi divergence
\citep{knoblauch2019generalized}. Alternatively
we could use another divergence measure in Eqn.~\ref{eqn:unregpred}---e.g.~the gamma divergence used
in Sec.~\ref{sec:robust}---to control the
qualitative features of $p_{\rm data}(y | \bx)$ that we would like to capture in $p(y| \bx)$.
We leave the exploration of these and other variants to future work.

\section{Related Work}
\label{sec:related}

The use of pseudo-inputs and inducing point methods to scale-up Gaussian Process inference has spawned a large literature, especially
in the context of variational inference \citep{csato2002sparse,seeger2003fast,quinonero2005unifying,snelson2006sparse,titsias2009variational, hensman2013gaussian,
hensman2015scalable,cheng2017variational}. While variational inference remains the most popular inference algorithm in the scalable GP setting,
researchers have also explored different variants of Expectation Propagation \citep{hernandez2016scalable,bui2017unifying} as well as
Stochastic gradient Hamiltonian Monte Carlo \citep{havasi2018inference} and other
MCMC algorithms \citep{hensman2015mcmc}.
For a recent review of scalable methods for GP
inference we refer the reader to \citep{liu2018gaussian}.

Our approach bears some resemblance to that described in \cite{raissi2019parametric} in that, like them, we consider
GP regression models that are parametric in nature. There are important differences, however. First because \citet{raissi2019parametric}
consider a two-step training procedure that does not benefit from a single coherent objective, they are unable to learn inducing point locations $\bZ$.
Second, inconsistent treatment of latent function uncertainty between the two training steps degrades the quality of the predictive uncertainty. For these reasons we find that the approach in \cite{raissi2019parametric} significantly underperforms\footnote{Both in terms of RMSE and log likelihood (especially the latter, with the predictive uncertainty drastically underestimated in some cases). See Sec.~\ref{sec:additional} for details.} all
the other methods we consider and hence we do not consider it further.

Several of our objective functions can also be seen as instances of Direct Loss Minimization, which emerges from a view of approximate inference as regularized loss minimization \cite{sheth2016monte,sheth2017excess,sheth2019pseudo}.\footnote{Note, however, that this interpretation is not applicable to our best performing
model class, PPGPR-MFD, which benefits from mean and variance functions that are decoupled.}
We refer the reader to Sec.~\ref{sec:dlm} in the supplementary materials for further discussion of this important connection.

Our focus on the predictive distribution also recalls \cite{snelson2005compact}, in which the authors construct parsimonious approximations
to Bayesian predictive distributions. Their approach differs from the approach adopted here, since the posterior distribution
is still computed (or approximated) as an intermediate step, whereas in PPGPR
we completely bypass the posterior. Similarly PPGPR recalls \cite{gordon2018meta}, where the authors consider training objectives that explicitly target posterior predictive distributions in the context of meta-learning.

\section{Experiments}
\label{sec:exp}
\begin{figure*}[t!]
  \centering
  \includegraphics[width=1.0\textwidth,center]{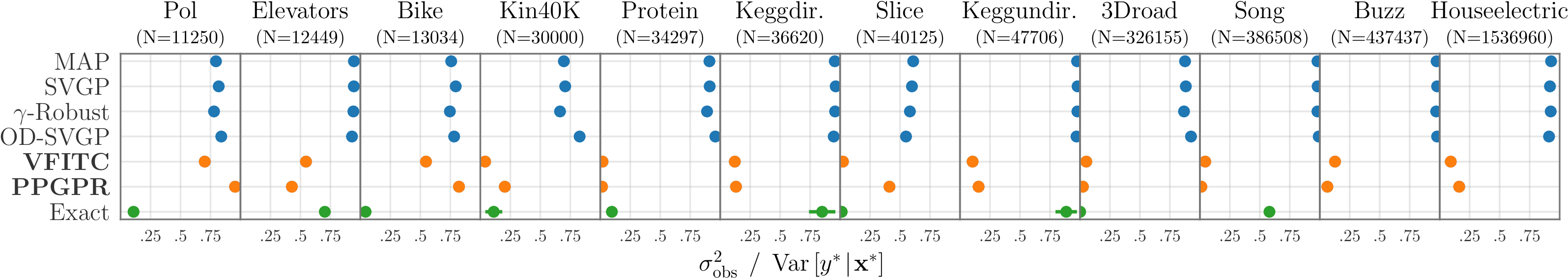}
    \caption{We depict the mean fraction of the predictive variance that is due to the observation noise (as measured on the test set). Results are averaged over ten random train/test/validation splits. See Sec.~\ref{sec:uci} for details.}
  \label{fig:ucivarratio}
\end{figure*}

In this section we compare the empirical performance of the approaches to scalable GP regression
introduced in Sec.~\ref{sec:paragp} to the baseline inference strategies
described in Sec.~\ref{sec:inference}.
All our models use a prior mean of zero and a Mat{\'e}rn kernel
with independent length scales for each input dimension.\footnote{For an implementation of our method in GPyTorch
see \texttt{https://git.io/JJy9b}.}

\subsection{Univariate regression}
\label{sec:uci}

We consider a mix of univariate regression datasets from the UCI repository \citep{Dua:2019},
with the number of datapoints ranging from $N\sim10^4$ to $N\sim10^6$ and the number of
input dimensions in the range ${\rm dim}(\bx) \in [3, 380]$. Among the methods introduced
in Sec.~\ref{sec:paragp}, we focus on Variational FITC
(Sec.~\ref{sec:vfitc}) and PPGPR-MFD (Sec.~\ref{sec:variants}). In addition to the baselines
reviewed in Sec.~\ref{sec:inference}, we also compare to the orthogonal parameterization of the basis decoupling method described in
\citet{cheng2017variational} and \citet{salimbeni2018orthogonally}, which we refer to as {\bf OD-SVGP}. For all
but the two largest datasets we also compare to {\bf Exact} GP inference, leveraging
the conjugate gradient approach described in \citep{wang2019exact}.

We summarize the results in Fig.~\ref{fig:ucill}-\ref{fig:ucivarratio} and Table~\ref{tab:ranks} (see the supplementary materials for additional results).
Both our approaches yield consistently lower negative log likelihoods (NLLs) than the baseline approaches,
with PPGPR performing particularly well. Averaged across all twelve datasets,
PPGPR outperforms the strongest baseline, OD-SVGP, by $\sim0.35$ nats.
Interestingly on most datasets PPGPR outperforms Exact GP inference.
We hypothesize that this is at least partially due to the ability
of our parametric models to encode heteroscedasticity, a ``bonus'' feature that was already noted by
\citet{snelson2006sparse}. Perhaps surprisingly, we note that on most datasets MAP
yields comparable NLLs to SVGP. Indeed averaged across all twelve datasets,
SVGP only outperforms MAP by $\sim 0.05$ nats.

The results for root mean squared errors (RMSE) exhibit somewhat less variability (see Fig.~\ref{fig:ucirmse}),
with PPGPR performing the best and OD-SVGP performing second best among the scalable methods.
In particular, in aggregate PPGPR attains the lowest RMSE (see Table~\ref{tab:ranks}), 
though it is outperformed by other methods on 2/12 datasets.
We hypothesize that the RMSE performance of VFITC could be substantially improved if the variational
distribution took the structured form that is implicitly used in OD-SVGP.
Not surprisingly Exact GP inference yields the lowest RMSE for most datasets.

Strikingly, both VFITC and PPGPR yield predictive variances that are in a qualitatively different
regime than those resulting from the scalable inference baselines.
Fig.~\ref{fig:ucivarratio} depicts the fraction of the total predictive variance $\text{Var} [ y^* \! \mid \! \bx ]$ due to the observation noise $\sigma_\text{obs}^2$.
The predictive variances from the baseline methods make relatively little use of input-dependent function uncertainty, instead relying primarily on the observation noise.
By contrast the variances of our parametric GP regressors are dominated by function uncertainty.
This substantiates the discussion in Sec.~\ref{sec:preddisc}.
Additionally, this observation explains the similar log likelihoods exhibited by SVGP and MAP: since neither method makes much use of function
uncertainty, the uncertainty encoded in $q(\bu)$ is of secondary importance.\footnote{Note that
MAP can be viewed as a degenerate limit of SVGP in which $q(\bu)$ is a Dirac delta function.}

Finally we note that for most datasets PPGPR prefers small values of $\betareg$. This
suggests that choosing $M \ll N$ is sufficient for ensuring well-regularized models and that overfitting is
not much of a concern in practice.

\begin{table}[t!]
  \centering
    \caption{Average ranking of methods (lower is better).
    CRPS is the Continuous Ranking Probability Score, a popular calibration metric for regression \citep{gneiting2007strictly}. Rankings are averages across datasets and splits. See Sec.~\ref{sec:uci} for details.}
  \label{tab:ranks}
  \resizebox{1.0\linewidth}{!}{%
    \begin{tabular}{cccccccc}
\toprule
{} &   &     MAP & $\gamma$-Robust &    SVGP & OD-SVGP &   VFITC &              PPGPR \\
\midrule
NLL  &   &  $5.47$ &          $4.76$ &  $4.33$ &  $3.08$ &  $2.20$ &  $\mathbf{ 1.17 }$ \\
RMSE &   &  $4.53$ &          $4.59$ &  $2.92$ &  $2.49$ &  $4.35$ &  $\mathbf{ 2.12 }$ \\
CRPS &   &  $5.55$ &          $4.12$ &  $4.21$ &  $3.20$ &  $2.90$ &  $\mathbf{ 1.02 }$ \\
\bottomrule
\end{tabular}

  }
\end{table}

\subsection{PPGPR Ablation Study}
\label{sec:ablation}

\begin{figure*}[t!]
  \centering
  \includegraphics[width=1.0\textwidth]{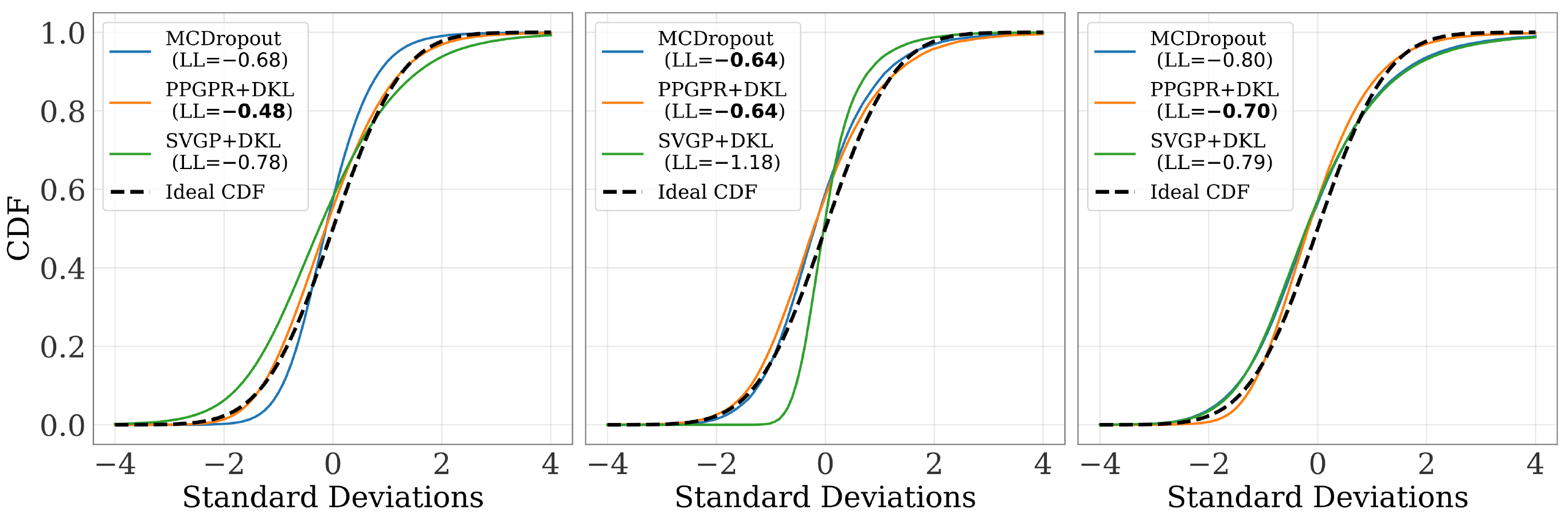}
    \caption{We visualize the calibration of three probabilistic deep learning models fit to trip data in three cities. For each method we depict the empirical CDF of the z-scores $z=(y^*-\mu_\bff(\bx^*))/\sigma(\bx^*)$ computed on the
    test set $\{(\bx^*_k, y^*_k)\}$.
    The ``Ideal CDF'' is the Normal CDF, which corresponds to the best possible calibration for a model with a Normal predictive distribution. See Sec.~\ref{sec:dkl} for details.}
  \label{fig:eta}
\end{figure*}

Encouraged by the predictive performance of PPGPR-MFD in Sec.~\ref{sec:uci}, we perform
a detailed comparison of the different PPGPR variants introduced in Sec.~\ref{sec:variants}.
Our results are summarized in Table \ref{tab:ranks_ablation} (see supplementary materials for additional figures).
We find that as we increase the capacity of the variance function $\sigma_\bff(\bx)^2$
(i.e.~PPGPR-$\delta$ $\Rightarrow$ PPGPR-MF $\Rightarrow$ PPGPR-Chol) the test NLL tends to decrease, while
the test RMSE tends to increase. This make sense since PPGPR prefers large predictive uncertainty in regions of input space where good data fit is hard to achieve. Consequently, there is less incentive to move the predictive mean function away from the prior in those regions, which can then result in higher RMSEs; this effect becomes more pronounced as the variance function becomes more flexible.
In contrast, since the mean and variance functions
are entirely decoupled in the case of PPGPR-MFD (apart from shared kernel hyperparameters), this model class obtains low NLLs without sacrificing performance on RMSE.

Note, finally, that PPGPR-$\delta$ and PPGPR-Chol utilize the same family of predictive distributions as MAP and SVGP, respectively,
and only differ in the objective function used during training.
If we compare these pairs of methods in Table \ref{tab:ranks_ablation}, we find that
PPGPR-$\delta$ and PPGPR-Chol yield the best log likelihoods, with PPGPR-Chol exhibiting degraded RMSE performance for the reason described
in the previous paragraph. Indeed these observations motivated the introduction of PPGPR-MFD.

\begin{table}[t!]
  \centering
  \caption{Average rank of PPGPR variants (lower is better).}
  \label{tab:ranks_ablation}
  \resizebox{1.0\linewidth}{!}{%
    \begin{tabular}{cccc|cccc}
\toprule
{} &   & \thead{ MAP } & \thead{ SVGP } & \thead{ PPGPR \\ ${\delta}$ } & \thead{ PPGPR \\ MF } & \thead{ PPGPR \\ Chol } & \thead{ PPGPR \\ MFD } \\
\midrule
NLL  &   &        $5.90$ &         $5.08$ &                        $3.96$ &                $2.66$ &       $\mathbf{ 1.59 }$ &                 $1.81$ \\
RMSE &   &        $3.64$ &         $2.36$ &                        $3.65$ &                $4.16$ &                  $5.38$ &      $\mathbf{ 1.82 }$ \\
CRPS &   &        $5.72$ &         $4.78$ &                        $3.87$ &                $2.77$ &                  $2.47$ &      $\mathbf{ 1.40 }$ \\
\bottomrule
\end{tabular}

  }
\end{table}

\subsection{Calibration in DKL Regression}
\label{sec:dkl}

In this section we demonstrate that PPGPR (introduced in Sec.~\ref{sec:mlpred}) offers an effective mechanism
for calibrating deep neural networks for regression.
To evaluate the potential of this approach, we utilize a real-world dataset of vehicular trip durations in three large cities.
In this setting, uncertainty estimation is critical for managing risk when estimating transportation costs.

We compare three methods: i) deep kernel learning \citep{wilson2016deep} using SVGP (SVGP+DKL);
ii) deep kernel learning using PPGPR-MFD (PPGPR+DKL); and
iii) MCDropout \citep{gal2016dropout}, a popular method for calibrating neural networks that does not rely
on Gaussian Processes.

In Fig.~\ref{fig:eta} we visualize how well each of the three methods is calibrated as compared to the best possible calibration for a model with Normal predictive distributions. Overall PPGPR+DKL performs the best,
with MCDropout outperforming or matching SVGP+DKL.
Using PPGPR has a number of additional advantages over MCDropout. In particular, because the predictive variances can be computed analytically for Gaussian Process models, the PPGPR+DKL model is significantly faster at test time than MCDropout, which requires forwarding data points through many sampled models (here 50). This is impractically slow for many applied settings, especially for large neural networks.


\section{Conclusions}
\label{sec:disc}

Gaussian Process regression with a Normal likelihood represents a peculiar case in that: i)
we can give an analytic formula for the exact posterior predictive distribution; but ii) it is impractical to compute for large datasets.
In this work we have argued that if our goal is high quality predictive distributions, it is sensible to bypass posterior
approximations and directly target the quantity of interest.
While this may be a bad strategy for an arbitrary probabilistic model, in
the case of GP regression inducing point methods provide a natural family of
parametric predictive distributions whose capacity can be controlled to prevent overfitting.
As we have shown empirically, the resulting predictive distributions exhibit significantly better calibrated uncertainties and higher log likelihoods than those obtained with other methods, which tend to yield overconfident uncertainty estimates that make little use of the kernel.

More broadly, our empirical results suggest that the good predictive performance of approximate GP methods like
SVGP may have more to do with the power of kernel methods than adherence to Bayes. The approach we have adopted here can
be viewed as maximum likelihood estimation for a model with a high-powered likelihood. This likelihood leverages flexible mean and variance functions
whose parametric form is motivated by inducing point methods. We suspect that a good ansatz for the variance function is of particular importance
for good predictive uncertainty. We explore one possible alternative---and much more flexible---parameterization in \citet{jankowiak2020deep}.
Here we suggest that one simple
and natural variant  of PPGPR would replace the mean function with a flexible neural network so that inducing points are only used to define the variance function.

\subsubsection*{Acknowledgements}
 MJ would like to thank Jeremias Knoblauch, Jack Jewson, and Theodoros Damouls for engaging conversations about
 robust variational inference. MJ would also like to thank Felipe Petroski Such for help with Uber compute infrastructure.
 This work was completed while MJ and JG were at Uber AI.


\bibliography{references}

\clearpage
\appendix
\section{Details on PPGPR-MFD}
\label{sec:suppmfd}

The regularizer we use for PPGPR-MFD is given by
\begin{equation}
\begin{split}
    \LL_{\rm reg} = \, \betareg \Big\{ & \log \NN(\bmm, \bK^{\mu})  - {\rm KL}(\NN(\bm{0}, \bS) \mid \NN(\bm{0}, \bK^{\sigma})) \Big \}\\
    \rightarrow \,-\tfrac{\betareg}{2} \Big\{ & {\rm Tr\,} \bS (\bK^{\sigma})^{-1} + \bmm^{\rm T} (\bK^{\mu})^{-1} \bmm \,+
                             \\ &\log \det \bK^{\sigma}  + \log \det \bK^{\mu} - \log \det \bS \Big\}
\end{split}
\end{equation}
with $\bK^{\sigma} \equiv \bK(\bZ_{\sigma}, \bZ_{\sigma})$ and
$\bK^{\mu}    \equiv \bK(\bZ_{\mu}, \bZ_{\mu})$, and where we have dropped irrelevant constants.
Here $\bZ_{\mu}$ and $\bZ_{\sigma}$ are the decoupled inducing point locations used to define
the mean function and variance function, respectively.
This regularizer can be viewed as the sum of two multivariate Normal KL divergences, one of which involves
a Dirac delta distribution (with divergent terms dropped).

As discussed in Sec.~\ref{sec:variants}, the mean and variance functions for PPGPR-MFD are given by
\begin{equation}
    \label{eqn:mfdmean}
    \mu_\bff(\bx_i) = \bk_i^{\mu T} (\bK^{\mu})^{-1} \bmm
\end{equation}
and
\begin{equation}
    \label{eqn:mfdvar}
    \sigma_\bff(\bx_i)^2 = \bKt^{\sigma}_{ii} + \bk_i^{\sigma T}  (\bK^{\sigma})^{-1} \bS (\bK^{\sigma})^{-1} \bk_i^{\sigma}
\end{equation}
where the various kernels in Eqn.~\ref{eqn:mfdmean} and Eqn.~\ref{eqn:mfdvar} share the same hyperparameters and $\bS$ is diagonal.

\section{Time and Space Complexity}
\label{sec:complexity}

We briefly describe the time and space complexity of the main algorithms for scalable GP regression
discussed in this work.
We note that our complexity analysis is similar to that of most sparse Gaussian process methods, such as SVGP \cite{hensman2013gaussian}.

\paragraph{Training complexity.}
Training a VFITC model requires optimizing Eqn.~\ref{eqn:vfitcelbo},
and training PPGPR models requires optimizing Eqn.~\ref{eqn:regpred}.
First, we note that the $\KL(q(\bu) | p(\bu))$ term is the $KL$ divergence between two multivariate Gaussians, which requires $\mathcal O(m^3)$ time complexity and $\mathcal O(m^2)$ space complexity.
For the remaining terms, the main computational bottleneck in both of these equations is computing the term $\bK_{MM}^{-1}$.
This is usually accomplished by computing its Cholesky factor $\bL$, which takes $\mathcal O(m^3)$ computation given $m$ inducing points.
After the Cholesky factor has been computed, all matrix solves involving $\bK_{MM}^{-1}$ require only $\mathcal O(m^2)$ computation.
The three main terms in both objective functions ($\mu_\bff(\bx_i)$, $\bKt_{ii}$, and $\bk_i^{T} \bK_{MM}^{-1} \bS \bK_{MM}^{-1} \bk_i$)
each require a constant number of matrix solves for each data point.
All together, the total time complexity of each training iteration is therefore $\mathcal O(m^3 + b m^2)$, where $b$ is the number of data points in a minibatch.
The space complexity is $\mathcal O(m^2 + bm)$ --- the size of storing $\bK_{MM}$, its Cholesky factor, and all $\bk_i$ vectors.

\paragraph{Prediction complexity.}
The predictive distributions for VFITC and PPGPR are given by Eqn.~\ref{eqn:preddist2}.
Again, the terms $\mu_\bff(\bx_i)$ and $\sigma_\bff(\bx_i) = \bKt_{ii} + \bk_i^{T} \bK_{MM}^{-1} \bS \bK_{MM}^{-1} \bk_i$
require a constant number of matrix solves with $\bK_{MM}^{-1}$ for each $\bx_i$.
However, we can cache and re-use the Cholesky factor of $\bK_{MM}^{-1}$ for all predictive distribution computations, since (after training) we are not updating the inducing point locations or the kernel hyperparameters.
Therefore, each predictive distribution has a time complexity of $\mathcal O(m^2)$ after the one-time cost of computing/caching the Cholesky factor.
It is also worth noting that predictive means can be computed in $\mathcal O(m)$ time by caching the vector $\bK_{MM}^{-1} \mathbf m$ vector.
The space complexity is also $\mathcal O(m^2)$ (the size of the Cholesky factor).

Note that each of our proposed variants in Section~\ref{sec:variants} have the same computational complexity, as each variant simply modifies the form of $\bS$ which is not the computational bottleneck.
We do note that the MFD variant requires roughly double the amount of computation and storage, as we are storing/performing solves with two $\bK_{MM}$ matrices (one for the predictive means and one for the predictive variances).
Finally, we note that the whitening parameterization (see Appendix~\ref{sec:whitened}) has the same computaitonal/storage complexity, as it also requires computing/storing a Cholesky factor.

\section{Experimental Details}
\label{sec:suppexp}

We use zero mean functions and Mat{\'e}rn kernels with independent length scales for each input dimension throughout.
All models and experiments are implemented using the GPyTorch framework \citep{gardner2018gpytorch} and
the Pyro probabilistic programming language \citep{bingham2019pyro}.

\subsection{Univariate regression}
\label{sec:suppuci}

We use the Adam optimizer for optimization with an initial learning rate of $\ell=0.01$ that is progressively
decimated over the course of training \citep{kingma2014adam}. We use a mini-batch size
of $B=10^4$ for the Buzz, Song, 3droad and Houseelectric datasets and $B=10^3$ for all other datasets.
We train for 400 epochs except for the Houseelectric dataset where we train for 200 epochs. Except
for the Exact results, we do 10 train/test/validation splits on all datasets (always in the proportion
15:3:2, respectively). In particular for the Exact results we do 3 train/test/validation splits on
the smaller datasets and one split for the two largest (3Droad and Song). All datasets are standardized in both input and output space; thus a predictive
distribution concentrated at zero yields a root mean squared error of unity. We use $M=1000$ inducing points
initialized with kmeans. In the case of OD-SVGP and PPGPR-MFD we use $M=1000$ inducing points for the mean
and $M=1000$ inducing points for the (co)variance. We use the validation set to determine a small set of hyperparameters.
In particular for SVGP, OD-SVGP, and VFITC we search over $\betareg \in \{0.1, 0.3, 0.5, 1.0\}$. For $\gamma$-Robust
we search over $\{1.01, 1.03, 1.05, 1.07\}$ (with $\betareg=1$). For all PPGPR variants we search
over $\betareg \in \{0.01, 0.05, 0.2, 1.0\}$. For MAP we fix $\betareg=1$.

\subsection{PPGPR Ablation}
\label{sec:suppablation}

The experimental procedure for the results reported in Sec.~\ref{sec:ablation} follows the procedure described
in Sec.~\ref{sec:suppuci}.

\begin{figure*}[t!]
  \centering
  \includegraphics[width=1.1\textwidth,center]{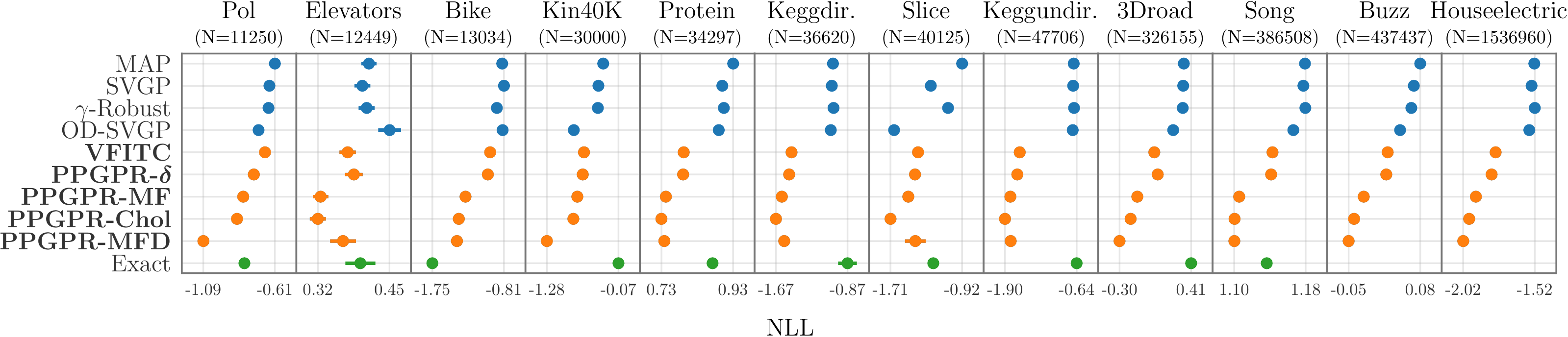}
    \caption{We compare test NLLs for the various methods explored in Sec.~\ref{sec:uci}-\ref{sec:ablation} in the main text
             (lower is better). Results are averaged over ten random train/test/validation splits. Here and throughout error bars depict standard errors.}
  \label{fig:ablation_ll}
\end{figure*}
\begin{figure*}[t!]
  \centering
  \includegraphics[width=1.1\textwidth,center]{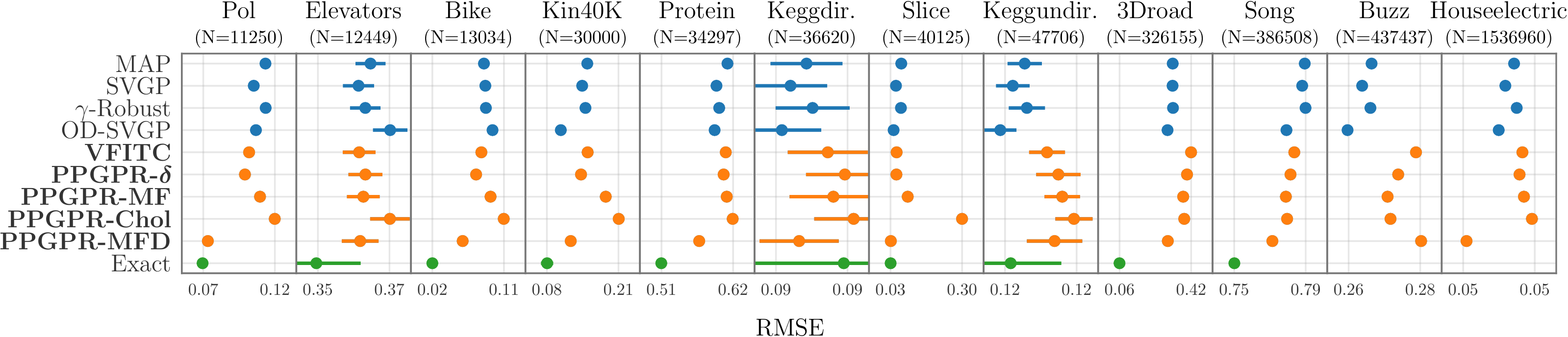}
    \caption{We compare test RMSEs for the various methods explored in Sec.~\ref{sec:uci}-\ref{sec:ablation} in the main text
             (lower is better). Results are averaged over ten random train/test/validation splits.}
  \label{fig:ablation_rmse}
\end{figure*}
\begin{figure*}[t!]
  \centering
  \includegraphics[width=1.1\textwidth,center]{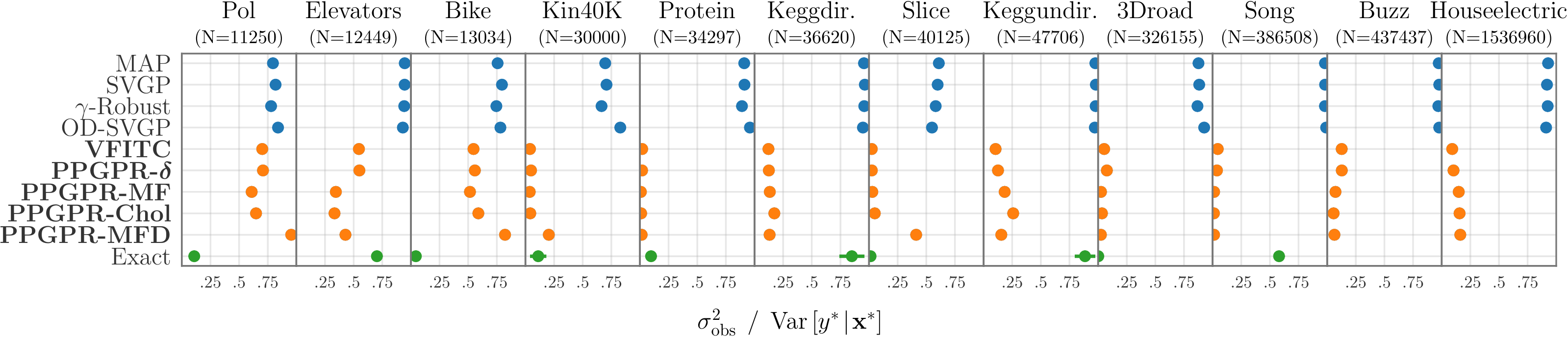}
    \caption{We compare the mean fraction of the predictive variance that is due to the observation noise for the various methods explored in Sec.~\ref{sec:uci}-\ref{sec:ablation} in the main text. Results are averaged over ten random train/test/validation splits.}
  \label{fig:ablation_var}
\end{figure*}
\begin{table*}[t!]
  \centering
  \caption{
    A compilation of all the results from Sec.~\ref{sec:uci}-\ref{sec:ablation}.
    For each metric and dataset we bold the result for the best performing scalable method.
    Errors are standard errors.
  }
  \label{tab:all_results}
  \resizebox{1.0\linewidth}{!}{%
    \begin{tabular}{cccccccccccccccc}
\toprule
     &               &  &                           MAP &                          SVGP &               $\gamma$-Robust &                       OD-SVGP &                         VFITC &  &           PPGPR-$\bm{\delta}$ &                      PPGPR-MF &                     PPGPR-Chol &                      PPGPR-MFD &  &               Exact \\
Metric & Dataset &          &                               &                               &                               &                               &                               &          &                               &                               &                                &                                &          &                     \\
\midrule
\midrule
NLL & Pol &          &            $-0.615 \pm 0.007$ &            $-0.651 \pm 0.005$ &            $-0.657 \pm 0.007$ &            $-0.723 \pm 0.006$ &            $-0.681 \pm 0.005$ &          &            $-0.755 \pm 0.009$ &            $-0.825 \pm 0.005$ &             $-0.866 \pm 0.005$ &  $\mathbf{ -1.090 \pm 0.009 }$ &          &  $-0.817 \pm 0.001$ \\
     & Elevators &          &             $0.412 \pm 0.007$ &             $0.401 \pm 0.007$ &             $0.409 \pm 0.007$ &             $0.448 \pm 0.010$ &             $0.376 \pm 0.007$ &          &             $0.386 \pm 0.008$ &  $\mathbf{ 0.329 \pm 0.007 }$ &   $\mathbf{ 0.324 \pm 0.007 }$ &              $0.368 \pm 0.011$ &          &   $0.398 \pm 0.013$ \\
     & Bike &          &            $-0.830 \pm 0.010$ &            $-0.807 \pm 0.007$ &            $-0.901 \pm 0.013$ &            $-0.824 \pm 0.009$ &            $-0.989 \pm 0.009$ &          &            $-1.019 \pm 0.008$ &            $-1.314 \pm 0.007$ &             $-1.402 \pm 0.004$ &  $\mathbf{ -1.426 \pm 0.010 }$ &          &  $-1.750 \pm 0.014$ \\
     & Kin40K &          &            $-0.333 \pm 0.002$ &            $-0.414 \pm 0.002$ &            $-0.423 \pm 0.002$ &            $-0.830 \pm 0.004$ &            $-0.658 \pm 0.001$ &          &            $-0.678 \pm 0.002$ &            $-0.770 \pm 0.002$ &             $-0.837 \pm 0.002$ &  $\mathbf{ -1.284 \pm 0.005 }$ &          &  $-0.075 \pm 0.001$ \\
     & Protein &          &             $0.931 \pm 0.003$ &             $0.902 \pm 0.003$ &             $0.906 \pm 0.004$ &             $0.892 \pm 0.006$ &             $0.796 \pm 0.004$ &          &             $0.794 \pm 0.004$ &  $\mathbf{ 0.747 \pm 0.008 }$ &   $\mathbf{ 0.735 \pm 0.007 }$ &   $\mathbf{ 0.743 \pm 0.008 }$ &          &   $0.875 \pm 0.002$ \\
     & Keggdir. &          &            $-1.032 \pm 0.017$ &            $-1.045 \pm 0.017$ &            $-1.026 \pm 0.021$ &            $-1.057 \pm 0.018$ &            $-1.494 \pm 0.014$ &          &            $-1.520 \pm 0.013$ &            $-1.601 \pm 0.016$ &  $\mathbf{ -1.667 \pm 0.028 }$ &             $-1.575 \pm 0.015$ &          &  $-0.870 \pm 0.052$ \\
     & Slice &          &            $-0.921 \pm 0.002$ &            $-1.267 \pm 0.003$ &            $-1.076 \pm 0.006$ &            $-1.673 \pm 0.013$ &            $-1.409 \pm 0.004$ &          &            $-1.442 \pm 0.004$ &            $-1.516 \pm 0.004$ &  $\mathbf{ -1.713 \pm 0.004 }$ &             $-1.438 \pm 0.057$ &          &  $-1.240 \pm 0.004$ \\
     & Keggundir. &          &            $-0.694 \pm 0.006$ &            $-0.704 \pm 0.006$ &            $-0.688 \pm 0.007$ &            $-0.712 \pm 0.006$ &            $-1.643 \pm 0.016$ &          &            $-1.685 \pm 0.018$ &            $-1.807 \pm 0.018$ &  $\mathbf{ -1.901 \pm 0.021 }$ &             $-1.801 \pm 0.013$ &          &  $-0.643 \pm 0.008$ \\
     & 3Droad &          &             $0.336 \pm 0.002$ &             $0.330 \pm 0.002$ &             $0.325 \pm 0.001$ &             $0.231 \pm 0.014$ &             $0.045 \pm 0.003$ &          &             $0.079 \pm 0.004$ &            $-0.122 \pm 0.002$ &             $-0.188 \pm 0.002$ &  $\mathbf{ -0.297 \pm 0.003 }$ &          &             $0.408$ \\
     & Song &          &             $1.180 \pm 0.001$ &             $1.178 \pm 0.001$ &             $1.181 \pm 0.001$ &             $1.168 \pm 0.001$ &             $1.145 \pm 0.001$ &          &             $1.143 \pm 0.001$ &             $1.109 \pm 0.001$ &   $\mathbf{ 1.104 \pm 0.001 }$ &   $\mathbf{ 1.103 \pm 0.001 }$ &          &             $1.139$ \\
     & Buzz &          &             $0.080 \pm 0.002$ &             $0.069 \pm 0.001$ &             $0.064 \pm 0.001$ &             $0.044 \pm 0.002$ &             $0.022 \pm 0.002$ &          &             $0.020 \pm 0.002$ &            $-0.020 \pm 0.001$ &             $-0.037 \pm 0.001$ &  $\mathbf{ -0.047 \pm 0.001 }$ &          &                 --- \\
     & Houseelectric &          &            $-1.524 \pm 0.001$ &            $-1.543 \pm 0.001$ &            $-1.521 \pm 0.001$ &            $-1.559 \pm 0.002$ &            $-1.794 \pm 0.001$ &          &            $-1.822 \pm 0.001$ &            $-1.931 \pm 0.001$ &             $-1.978 \pm 0.001$ &  $\mathbf{ -2.020 \pm 0.003 }$ &          &                 --- \\
\midrule
RMSE & Pol &          &             $0.115 \pm 0.001$ &             $0.107 \pm 0.002$ &             $0.115 \pm 0.002$ &             $0.109 \pm 0.001$ &             $0.104 \pm 0.002$ &          &             $0.101 \pm 0.001$ &             $0.111 \pm 0.002$ &              $0.121 \pm 0.002$ &   $\mathbf{ 0.077 \pm 0.001 }$ &          &   $0.074 \pm 0.001$ \\
     & Elevators &          &  $\mathbf{ 0.364 \pm 0.002 }$ &  $\mathbf{ 0.360 \pm 0.003 }$ &  $\mathbf{ 0.362 \pm 0.002 }$ &             $0.370 \pm 0.003$ &  $\mathbf{ 0.360 \pm 0.003 }$ &          &  $\mathbf{ 0.362 \pm 0.003 }$ &  $\mathbf{ 0.362 \pm 0.003 }$ &              $0.370 \pm 0.003$ &   $\mathbf{ 0.361 \pm 0.003 }$ &          &   $0.347 \pm 0.007$ \\
     & Bike &          &             $0.086 \pm 0.002$ &             $0.088 \pm 0.002$ &             $0.089 \pm 0.002$ &             $0.097 \pm 0.002$ &             $0.083 \pm 0.002$ &          &             $0.076 \pm 0.001$ &             $0.094 \pm 0.002$ &              $0.111 \pm 0.002$ &   $\mathbf{ 0.060 \pm 0.001 }$ &          &   $0.022 \pm 0.001$ \\
     & Kin40K &          &             $0.156 \pm 0.001$ &             $0.147 \pm 0.001$ &             $0.152 \pm 0.001$ &  $\mathbf{ 0.109 \pm 0.001 }$ &             $0.156 \pm 0.001$ &          &             $0.145 \pm 0.001$ &             $0.188 \pm 0.002$ &              $0.211 \pm 0.003$ &              $0.126 \pm 0.001$ &          &   $0.084 \pm 0.005$ \\
     & Protein &          &             $0.610 \pm 0.002$ &             $0.594 \pm 0.002$ &             $0.598 \pm 0.002$ &             $0.591 \pm 0.003$ &             $0.607 \pm 0.002$ &          &             $0.604 \pm 0.002$ &             $0.609 \pm 0.002$ &              $0.618 \pm 0.002$ &   $\mathbf{ 0.569 \pm 0.002 }$ &          &   $0.514 \pm 0.003$ \\
     & Keggdir. &          &  $\mathbf{ 0.087 \pm 0.001 }$ &  $\mathbf{ 0.086 \pm 0.001 }$ &  $\mathbf{ 0.088 \pm 0.001 }$ &  $\mathbf{ 0.085 \pm 0.001 }$ &             $0.089 \pm 0.001$ &          &             $0.090 \pm 0.001$ &             $0.089 \pm 0.002$ &              $0.090 \pm 0.001$ &   $\mathbf{ 0.087 \pm 0.001 }$ &          &   $0.090 \pm 0.004$ \\
     & Slice &          &             $0.071 \pm 0.001$ &             $0.051 \pm 0.001$ &             $0.070 \pm 0.003$ &             $0.043 \pm 0.001$ &             $0.054 \pm 0.001$ &          &             $0.053 \pm 0.001$ &             $0.095 \pm 0.001$ &              $0.298 \pm 0.004$ &   $\mathbf{ 0.032 \pm 0.001 }$ &          &   $0.031 \pm 0.003$ \\
     & Keggundir. &          &             $0.121 \pm 0.001$ &  $\mathbf{ 0.120 \pm 0.001 }$ &             $0.121 \pm 0.001$ &  $\mathbf{ 0.119 \pm 0.001 }$ &             $0.123 \pm 0.001$ &          &             $0.123 \pm 0.001$ &             $0.124 \pm 0.001$ &              $0.125 \pm 0.001$ &              $0.123 \pm 0.001$ &          &   $0.119 \pm 0.002$ \\
     & 3Droad &          &             $0.329 \pm 0.001$ &             $0.329 \pm 0.001$ &             $0.331 \pm 0.001$ &  $\mathbf{ 0.303 \pm 0.004 }$ &             $0.424 \pm 0.001$ &          &             $0.403 \pm 0.005$ &             $0.383 \pm 0.002$ &              $0.389 \pm 0.001$ &   $\mathbf{ 0.304 \pm 0.001 }$ &          &             $0.057$ \\
     & Song &          &             $0.788 \pm 0.001$ &             $0.786 \pm 0.001$ &             $0.788 \pm 0.001$ &             $0.778 \pm 0.001$ &             $0.782 \pm 0.001$ &          &             $0.780 \pm 0.001$ &             $0.777 \pm 0.001$ &              $0.778 \pm 0.001$ &   $\mathbf{ 0.770 \pm 0.001 }$ &          &             $0.750$ \\
     & Buzz &          &             $0.265 \pm 0.001$ &             $0.261 \pm 0.000$ &             $0.264 \pm 0.001$ &  $\mathbf{ 0.256 \pm 0.001 }$ &             $0.281 \pm 0.001$ &          &             $0.275 \pm 0.001$ &             $0.271 \pm 0.001$ &              $0.272 \pm 0.001$ &              $0.283 \pm 0.001$ &          &                 --- \\
     & Houseelectric &          &             $0.052 \pm 0.000$ &             $0.051 \pm 0.000$ &             $0.052 \pm 0.000$ &             $0.050 \pm 0.000$ &             $0.053 \pm 0.000$ &          &             $0.052 \pm 0.000$ &             $0.053 \pm 0.000$ &              $0.054 \pm 0.000$ &   $\mathbf{ 0.046 \pm 0.000 }$ &          &                 --- \\
\midrule
CRPS & Pol &          &             $0.065 \pm 0.001$ &             $0.061 \pm 0.000$ &             $0.062 \pm 0.001$ &             $0.059 \pm 0.000$ &             $0.060 \pm 0.000$ &          &             $0.057 \pm 0.001$ &             $0.057 \pm 0.000$ &              $0.057 \pm 0.000$ &   $\mathbf{ 0.040 \pm 0.000 }$ &          &   $0.051 \pm 0.000$ \\
     & Elevators &          &             $0.200 \pm 0.001$ &             $0.198 \pm 0.001$ &             $0.199 \pm 0.001$ &             $0.203 \pm 0.001$ &             $0.197 \pm 0.001$ &          &             $0.198 \pm 0.001$ &  $\mathbf{ 0.193 \pm 0.001 }$ &   $\mathbf{ 0.195 \pm 0.001 }$ &   $\mathbf{ 0.195 \pm 0.002 }$ &          &   $0.195 \pm 0.003$ \\
     & Bike &          &             $0.047 \pm 0.000$ &             $0.049 \pm 0.000$ &             $0.043 \pm 0.000$ &             $0.051 \pm 0.001$ &             $0.042 \pm 0.000$ &          &             $0.040 \pm 0.000$ &             $0.037 \pm 0.000$ &              $0.037 \pm 0.001$ &   $\mathbf{ 0.028 \pm 0.000 }$ &          &   $0.019 \pm 0.000$ \\
     & Kin40K &          &             $0.088 \pm 0.000$ &             $0.082 \pm 0.000$ &             $0.082 \pm 0.000$ &             $0.056 \pm 0.000$ &             $0.074 \pm 0.000$ &          &             $0.071 \pm 0.000$ &             $0.077 \pm 0.000$ &              $0.079 \pm 0.000$ &   $\mathbf{ 0.050 \pm 0.000 }$ &          &   $0.093 \pm 0.000$ \\
     & Protein &          &             $0.337 \pm 0.001$ &             $0.326 \pm 0.001$ &             $0.326 \pm 0.001$ &             $0.317 \pm 0.001$ &             $0.318 \pm 0.001$ &          &             $0.317 \pm 0.001$ &             $0.310 \pm 0.001$ &              $0.310 \pm 0.001$ &   $\mathbf{ 0.288 \pm 0.001 }$ &          &   $0.293 \pm 0.001$ \\
     & Keggdir. &          &             $0.038 \pm 0.000$ &             $0.037 \pm 0.000$ &             $0.036 \pm 0.000$ &             $0.037 \pm 0.000$ &             $0.033 \pm 0.000$ &          &             $0.032 \pm 0.000$ &             $0.031 \pm 0.000$ &   $\mathbf{ 0.031 \pm 0.000 }$ &              $0.031 \pm 0.000$ &          &   $0.046 \pm 0.002$ \\
     & Slice &          &             $0.044 \pm 0.000$ &             $0.031 \pm 0.000$ &             $0.037 \pm 0.000$ &             $0.021 \pm 0.000$ &             $0.029 \pm 0.000$ &          &             $0.028 \pm 0.000$ &             $0.032 \pm 0.000$ &              $0.060 \pm 0.001$ &   $\mathbf{ 0.014 \pm 0.000 }$ &          &   $0.029 \pm 0.000$ \\
     & Keggundir. &          &             $0.052 \pm 0.000$ &             $0.051 \pm 0.000$ &             $0.050 \pm 0.000$ &             $0.051 \pm 0.000$ &             $0.038 \pm 0.000$ &          &             $0.037 \pm 0.000$ &             $0.036 \pm 0.000$ &   $\mathbf{ 0.035 \pm 0.000 }$ &              $0.036 \pm 0.000$ &          &   $0.056 \pm 0.001$ \\
     & 3Droad &          &             $0.178 \pm 0.000$ &             $0.177 \pm 0.000$ &             $0.175 \pm 0.000$ &             $0.163 \pm 0.002$ &             $0.191 \pm 0.001$ &          &             $0.185 \pm 0.002$ &             $0.170 \pm 0.001$ &              $0.167 \pm 0.000$ &   $\mathbf{ 0.138 \pm 0.000 }$ &          &             $0.142$ \\
     & Song &          &             $0.441 \pm 0.000$ &             $0.440 \pm 0.000$ &             $0.441 \pm 0.000$ &             $0.434 \pm 0.000$ &             $0.434 \pm 0.000$ &          &             $0.433 \pm 0.000$ &             $0.426 \pm 0.000$ &              $0.425 \pm 0.000$ &   $\mathbf{ 0.422 \pm 0.000 }$ &          &             $0.420$ \\
     & Buzz &          &             $0.141 \pm 0.000$ &             $0.140 \pm 0.000$ &             $0.139 \pm 0.000$ &             $0.135 \pm 0.000$ &             $0.141 \pm 0.000$ &          &             $0.140 \pm 0.000$ &             $0.136 \pm 0.000$ &              $0.135 \pm 0.000$ &   $\mathbf{ 0.134 \pm 0.000 }$ &          &                 --- \\
     & Houseelectric &          &             $0.027 \pm 0.000$ &             $0.027 \pm 0.000$ &             $0.027 \pm 0.000$ &             $0.026 \pm 0.000$ &             $0.025 \pm 0.000$ &          &             $0.025 \pm 0.000$ &             $0.024 \pm 0.000$ &              $0.024 \pm 0.000$ &   $\mathbf{ 0.022 \pm 0.000 }$ &          &                 --- \\
\bottomrule
\end{tabular}

  }
\end{table*}

\subsection{DKL calibration}
\label{sec:suppdkl}

MC-Dropout has two hyperparameters that must be set by hand: a dropout proportion $p$ and a prior variance inflation term
$\tau$. These were set by temporarily removing a portion of the training data as a validation set and performing a small
grid search on each dataset over $p \in [0.05, 0.1, 0.15, 0.2, 0.25]$ and $\tau \in [0.05, 0.1, 0.25, 0.5, 0.75, 1.0]$.
For PPGPR-MFD, $\beta \in [0.05, 0.2, 0.5, 1.0]$ was chosen in a similar fashion. The datasets for each of
the three cities contain 30 features that encode various aspects of a trip like origin and destination location, time of
day and week, as well as various rudimentary routing features.

For all three methods, we use the same five layer fully connected neural network, with hidden representation sizes of $
[256, 256, 128, 128, 64]$ and ReLU nonlinearities. We use the Adam optimizer and use an
initial learning rate of $\ell=0.01$, which we drop by a factor of $0.1$ at $100$ and $150$ epochs.
We train for 200 epochs for all three methods. For the GP methods, we use $M=1024$ inducing points,
initialized by randomly selecting training data points and passing them through the initial feature extractor.

\begin{figure*}[t!]
  \centering
  \includegraphics[width=1.1\textwidth,center]{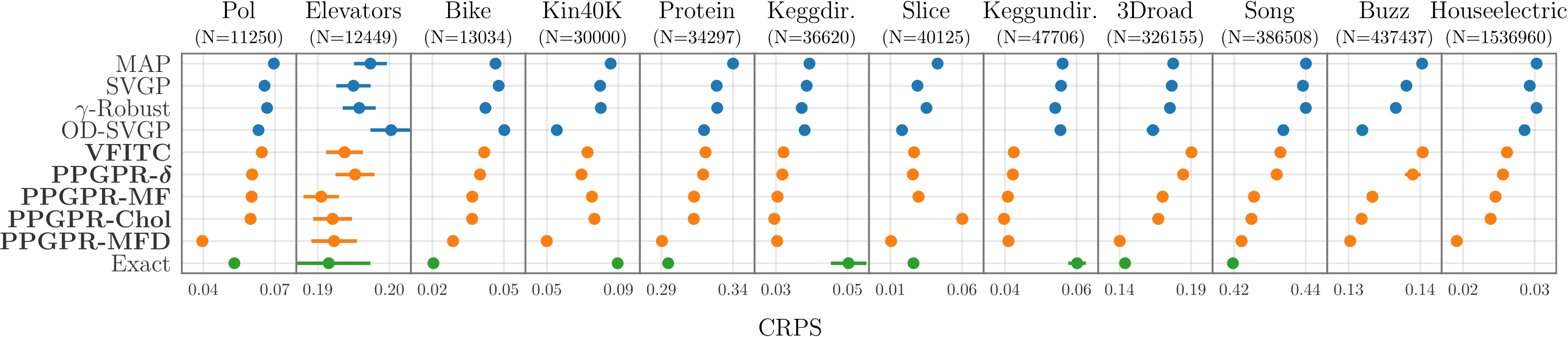}
  \vspace{-2em}
    \caption{We compare test continuous ranked probability scores (CRPS) for the various methods explored in Sec.~\ref{sec:uci}-\ref{sec:ablation} (lower is better). Results are averaged over ten random train/test/validation splits.}
  \label{fig:ucicrps}
\end{figure*}

\section{Additional Experimental Results}
\label{sec:additional}

In Fig.~\ref{fig:ablation_ll}-\ref{fig:ucicrps} we depict summary results for all the experiments
in Sec.~\ref{sec:uci}-\ref{sec:ablation} in the main text.
In particular in Fig.~\ref{fig:ucicrps} we depict Continuous Ranked Probability Scores \citep{gneiting2007strictly}.
We also include a complete compilation of our results in Table \ref{tab:all_results}.

\paragraph{The effect of $\betareg$}

We find empirically that PPGPR is robust to the value of $\betareg$. On the five smallest UCI datasets, the test RMSE varies by no more than 5\% (relative) and the LL by no more than 0.05 nats as we vary $\betareg$ from 0 to 1. This is not unexpected, since we are always in the regime $M \ll N$.

\paragraph{Comparison to Raissi et.~al.}

We do not include a full comparison to the method in \citet{raissi2019parametric} because we find that it is outperformed by all the
other baselines we consider. In particular while this method can achieve middling RMSEs, on many datasets it achieves very poor log likelihoods. For example, using the authors' implementation\footnote{\texttt{https://github.com/maziarraissi/ParametricGP}} we find
a test NLL of $\sim26$ nats on the Elevators dataset and $\sim6$ nats on the Pol dataset. This performance can be traced to the incoherency of the two-objective approach adopted by this method. In particular the logic of the derivation makes it unclear whether the noise term in the kernel should be included during test time; this ambiguity is reflected in the authors' code,\footnote{See line 177 in \texttt{parametric\char`_GP.py}.} where this contribution is by default commented out. While the NLL performance can be improved by including this term, the larger point is that this approach does not provide a coherent account of function uncertainty.

\paragraph{In-Sample Comparison to Exact GPs}

\begin{figure*}[t!]
  \centering
  \includegraphics[width=1.1\textwidth,center]{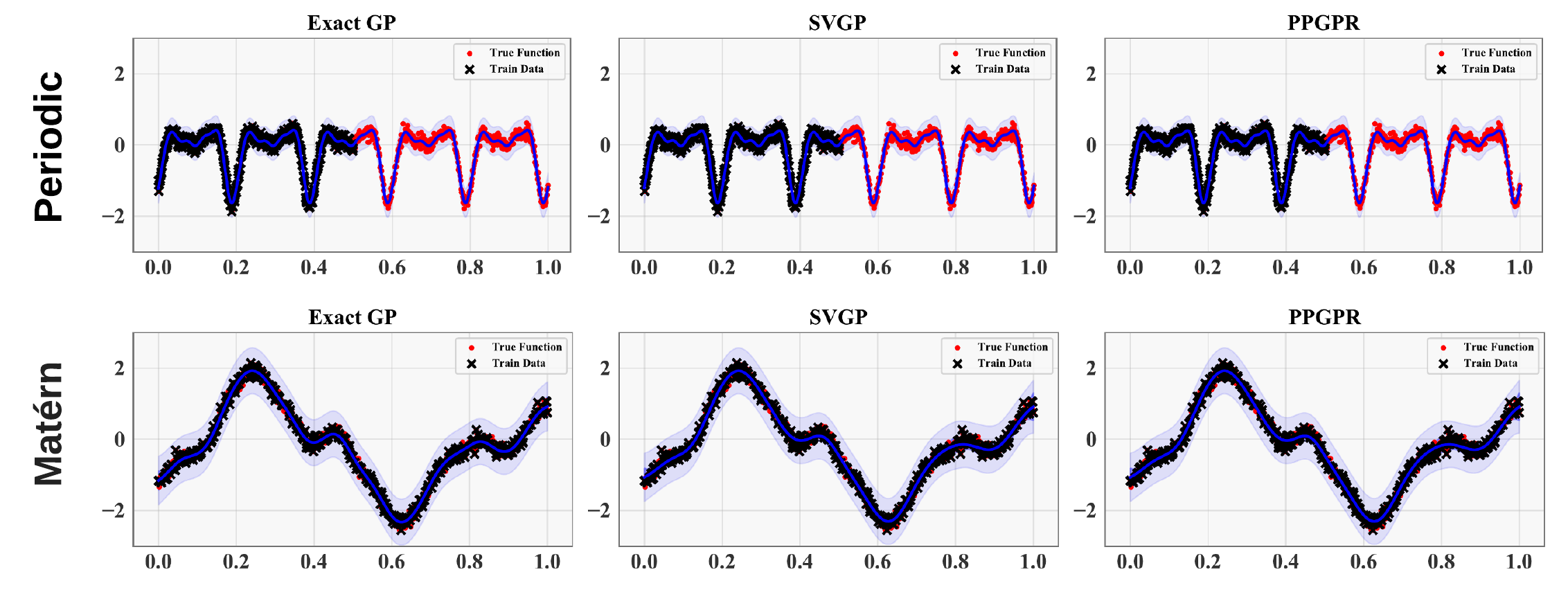}
  \vspace{-2em}
    \caption{We compare exact GPs, SVGP, and PPGPR from functions drawn from a Gaussian process prior using a periodic kernel (top) and Matern kernel (bottom).}
  \label{fig:exact_gp_in_sample}
\end{figure*}

In this section, we compare the fit of an exact GP, of SVGP, and of PPGPR to samples drawn from a Gaussian process prior with a Matern kernel and a Periodic kernel. For the draw from the Matern kernel,
we use a lengthscale of 0.1 and an outputscale of 1. For the periodic kernel, we use a period length of 0.2 and outputscale of 1. For both kernels, we draw the function on the range $[0, 1]$, and fit all three models starting from default hyperparameter initializations. For the periodic
kernel case, we consider an extrapolation task. The results are presented in Fig~\ref{fig:exact_gp_in_sample}.
We observe that all three methods are capable of performing the extrapolation task using the periodic kernel, and in general depict qualitatively similar results. This verifies that--in very simple cases--much of the model performance
can depend on the choice of kernel rather than the particular training scheme.
\section{Whitened Sparse Gaussian Process Regression}
\label{sec:whitened}

The hyperparameters and variational parameters of the models can be learned by directly optimizing the objective functions in Eqn.~\ref{eqn:jensenenergy} (for MAP), in Eqn.~\ref{eqn:svgp} (for SVGP), in Eqn.~\ref{eqn:vfitcelbo} (for VFITC), and Eqn.~\ref{eqn:regpred} (for PPGPR).
In practice, we modify these objective functions using a
transformation proposed by \citep{matthews2017scalable}.
The ``whitening transformation'' is a simple change of variables
\[ \bu' = \bhalfK_{MM}^{-1} \bu \]
where $\bhalfK_{MM}$ is a matrix such that $\bhalfK_{MM} \bhalfK_{MM}^\top = \bK_{MM}$.
(Typically, $\bhalfK_{MM}$ is taken to be the Cholesky factor of $\bK_{MM}$.)
Intuitively, this transformation is advantageous because it reduces the number of changing terms in the objective functions.
In whitened coordinates the prior $p(\bu')$ is constant: $p(\bu') = \NN(\mathbf 0, \bhalfK_{MM}^{-1} \bK_{MM} \bhalfK_{MM}^{-\top}) = \NN(\mathbf 0, \mathbf{I})$.
Incorporating this transformation into Eqn.~\ref{eqn:jensenenergy} gives us
\begin{align*}
\begin{aligned}
&\log p(\by, \bu' |\bX, \bZ) \ge  \EE_{p(\bff|\bu')} \left[ \log p(\by|\bff) + \log p(\bu') \right] \\
  &=  \sum_{i=1}^n \log \mathcal{N}(y_i | \bk_i^{T} \bhalfK_{MM}^{-1} \bu', \sigma_\text{obs}) -  \tfrac{1}{2\sigma_\text{obs}}  {\rm Tr}\; \bKt_{NN} \\
  &\phantom{=} \:\: - \tfrac{1}{2} \Vert \bu' \Vert_2^2 - \tfrac{M}{2} \log(2 \pi).
 \end{aligned}
\end{align*}
Importantly, the prior term $p(\bu')$ in the modified objective does not depend on the inducing point locations $\bZ$ or kernel hyperparameters.
In all experiments we use similarly modified objectives for better optimization.
For MAP and PPGPR-$\delta$, we directly optimize the whitened variables $\bu'$.
For SVGP, VFITC, PPGPR-Chol, and PPGPR-MF, the whitened variational distribution is given as $q(\bu') = \NN(\bhalfK_{MM}^{-1} \bmm, \bhalfK_{MM}^{-1} \bS \bhalfK_{MM}^{-1})$.
We parameterize the mean with a vector $\bmm' = \bhalfK_{MM}^{-1} \bmm$ and we parameterize the covariance with a lower triangular matrix $\bL \bL^\top = \bhalfK_{MM}^{-1} \bS \bhalfK_{MM}^{-1}$.
The same transformation is applied to the OD-SVGP covariance variational parameters---referred to as the $\beta$ parameters by \citet{salimbeni2018orthogonally}.
As in the original work, the mean variational parameters (the $\gamma$ parameters) are not re-parameterized.

We apply a similar whitening transformation to the decoupled variant of our method (PGPR-MFD).
Given inducing points $\bZ_\mu$ and $\bZ_\sigma$ for the mean and variance functions respectively, we optimize the transformed parameters
\[ \bmm' = \bhalfK_{MM}^{(\mu)-1} \bmm, \quad \bS' = \bhalfK_{MM}^{(\sigma)-1} \bS \bhalfK_{MM}^{(\sigma)-\top}, \]
where $\bhalfK_{MM}^{(\mu)}$ and $\bhalfK_{MM}^{(\sigma)}$ are the Cholesky factor of the $\bZ_\mu$ and $\bZ_\sigma$ kernel matrices.
$\bS'$ is constrained to be a diagonal matrix because of the mean-field approximation.

\section{Connection to Direct Loss Minimization}
\label{sec:dlm}

In a series of papers the authors of \citep{sheth2016monte,sheth2017excess,sheth2019pseudo}
consider approximate inference from the angle of regularized loss minimization.
For Bayesian models of the general form
\begin{equation}
p(\by, \bu) = p(\bu) \prod_{i=1}^N p(y_i | \bu, \bx_i)
\end{equation}
and (approximate posterior) distributions $q(\bu)$
Sheth and Khardon investigate PAC-Bayes-like bounds for the Bayes risk
\begin{equation}
r_{\rm Bayes}[q(\bu)] \equiv  \EE_{p_{\rm data}(\bx, y)} \left[- \log \EE_{q(\bu)} p(y | \bu, \bx) \right]
\end{equation}
and the Gibbs risk
\begin{equation}
r_{\rm Gibbs}[q(\bu)] \equiv  \EE_{p_{\rm data}(\bx, y)} \EE_{q(\bu)} \left[- \log p(y | \bu, \bx) \right]
\end{equation}
In particular, under some restrictions on the family of multivariate Normal distributions $q(\bu)$,
they establish a bound on a smoothed log loss variant of Bayes risk for an inference setup
that corresponds to Variational FITC (Sec.~\ref{sec:vfitc}).\footnote{See corollary 12 in \citep{sheth2017excess}.}
Additionally, under similar assumptions an analogous bound for an inference setup that corresponds to
PPGPR-Chol is established in \citep{sheth2019pseudo}.
Moreover, in \citep{sheth2016monte} the authors provide some empirical evidence for the good performance
of these two objective functions.

Importantly, our parametric modeling perspective allows us more flexibility than is allowed by Direct Loss Minimization as
explored by Sheth and Khardon.
Indeed much of the predictive power of approximate GP regressors arguably comes from the sensible behavior of the parametric
family of mean and variance functions that is induced by a particular kernel and associated inducing point locations $\bZ$---for example, the variance $\sigma_{\bff}(\bx)^2$ increases as $\bx$ moves away from $\bZ$. In particular the maximum
likelihood approach we adopt in Sec.~\ref{sec:mlpred} allows us to consider parametric families of predictive
distributions that are \emph{not} of the form $\EE_{q(\bu)} \EE_{p(f|\bu,\bx)} \left[ p(y | f) \right]$
for some distribution $q(\bu)$. This is in fact the case for our best performing method, PPGPR-MFD, which
freely combines mean and variance functions that are parameterized by distinct inducing point locations
$\bZ_{\mu}$ and $\bZ_{\sigma}$. Indeed, while we have not done so in our experiments, this perspective
allows us to entirely decouple $\mu_{\bff}(\bx)$ and $\sigma_{\bff}(\bx)^2$ by introducing separate kernels for each.

\section{Connection to stochastic EP and the BB-$\alpha$ objective}
\label{sec:ep}

Suppose we want to approximate the distribution
\begin{equation}
\label{eqn:epsetup}
    p(\omega) = \frac{1}{Z} p_0(\omega) \prod_{i=1}^N f_n(\omega)
\end{equation}
where $Z$ is an unknown normalizer.
In the prototypical context of Bayesian modeling $p_0(\omega)$ would be a prior distribution
and $f_n(\omega)$ would be a likelihood factor for the $n^{\rm th}$ datapoint.
Expectation propagation (EP) is a broad class of algorithms that can be used to approximate
distributions like that in Eqn.~\ref{eqn:epsetup} \citep{minka2004power}. In the following we give a brief
review of a few variants of EP and describe a connection to the Predictive objective defined in
Eqn.~\ref{eqn:regpred} in the main text.

 \citet{li2015stochastic}
propose a particular variant of EP called Stochastic EP that reduces memory requirements
by a factor of $N$ by tying (i.e.~sharing) factors together.
In subsequent work \citet{hernandez2016black} present a version of Stochastic EP
that is formulated in terms of an energy function, the so-called
BB-$\alpha$ objective $\LL_{\alpha}$, which is given by
\begin{equation}
    \LL_{\alpha} = -\frac{1}{\alpha} \sum_{i=1}^N \log \EE_{q(\omega)} \left[
        \left( \frac{f_n(\omega) p_0(\omega)^{1/N}}{q(\omega)^{1/N}}\right)^{\alpha}
        \right]
\end{equation}
where $q(\omega)$ is a so-called cavity distribution and $\alpha \in [0,1]$.
As shown in \citep{li2017dropout}, this can be rewritten as
\begin{equation}
    \LL_{\alpha} = R_{\xi}(\qtilde || p_0) -
    \tfrac{1}{\alpha}\sum_{i=1}^N \log \EE_{\qtilde} \left[ f_n(\omega)^{\alpha} \right]
\end{equation}
where $\qtilde(\omega)$ is defined by the equation
\begin{equation}
q(\omega) = \frac{1}{Z_q} \qtilde(\omega) \left(\frac{\qtilde(\omega)}{p_0(\omega)} \right)^{\frac{\alpha}{N - \alpha }}
\end{equation}
and where $\xi \equiv \tfrac{N}{N-\alpha}$ and $R_{\xi}(\qtilde || p_0)$ is a R\'enyi divergence.
\citet{li2017dropout} then argue that, under
suitable conditions, we have that as $\tfrac{\alpha}{N} \rightarrow 0$ this becomes
\begin{equation}
    \LL_{\alpha} \rightarrow \KL(q || p_0) -
    \tfrac{1}{\alpha}\sum_{i=1}^N \log \EE_{q} \left[ f_n(\omega)^{\alpha} \right]
\end{equation}
For the particular choice $\alpha=1$ (so that we require $N\rightarrow \infty$) this then becomes
\begin{equation}
\label{eqn:alpha1}
    \LL_{\alpha=1} \rightarrow \KL(q || p_0) -
    \sum_{i=1}^N \log \EE_{q} \left[ f_n(\omega) \right]
\end{equation}
The similarity of Eqn.~\ref{eqn:alpha1} and Eqn.~\ref{eqn:regpred} is now manifest.

For further discussion of EP methods in the context of Gaussian Process inference
we refer the reader to \citep{bui2017unifying}.

\end{document}